\documentclass[letterpaper, 10 pt, conference]{ieeeconf}

\IEEEoverridecommandlockouts  

\usepackage{fix-cm}
\usepackage{etex}

\usepackage{dblfloatfix}

\usepackage{nag}

\makeatletter
\@ifpackageloaded{xcolor}{}{%
\usepackage[table,x11names,dvipsnames,svgnames]{xcolor}%
}
\makeatother

\usepackage{colortbl}

\usepackage{graphicx}
\usepackage{wrapfig}

\definecolorset{RGB}{lyft}{}{Red,194,39,36;Sunset,202,53,33;Orange,205,68,20;Amber,200,117,42;Yellow,242,169,52;Citron,186,188,44;Lime,112,159,33;Green,56,139,31;Mint,45,118,56;Teal,52,133,135;Cyan,60,132,202;Blue,55,94,248;Indigo,64,13,247;Purple,115,42,248;Pink,176,25,145;Rose,176,32,75}

\definecolorset{HTML}{h}{}{grapefruit,ED5565;bittersweet,FC6E51;sunflower,FFCE54;grass,A0D468;mint,48CFAD;aqua,4FC1E9;bluejeans,5D9CEC;lavender,AC92EC;pinkrose,EC87C0;lightgray,F5F7FA;mediumgray,CCD1D9;darkgray,656D78}

\usepackage{cite}

\usepackage{microtype}

\usepackage[american]{babel}

\usepackage{array}
\usepackage{multirow}
\usepackage{booktabs}
\usepackage{makecell} 

\ifcsname labelindent\endcsname
\let\labelindent\relax
\fi
\usepackage[inline]{enumitem}

\usepackage{subfig}

\newenvironment{lenumerate}[2][]
{\begin{enumerate}[label=(#2\arabic*),leftmargin=0.2in,itemindent=0.15in,#1]}
{\end{enumerate}}

\setlist*[enumerate,1]{label={\itshape\arabic*)}}

\makeatletter
\newcommand{\paragraphswithstop}{%
\let\copyparagraph\paragraph%
\renewcommand\paragraph[1]{\copyparagraph{##1.}}%
}
\makeatother

\usepackage[framemethod=tikz]{mdframed}

\makeatletter
\def\namedlabel#1#2{\begingroup
  #2%
  \def\@currentlabel{#2}%
  \phantomsection\label{#1}\endgroup
}
\makeatother

\makeatletter
\def\namedlabelphantom#1#2{\begingroup
  \def\@currentlabel{#2}%
  \phantomsection\label{#1}\endgroup
}
\makeatother

\newcommand{\parunskip}{\bgroup\unskip\parfillskip=0pt \par\egroup}

\input{math}
\newcommand{\real}[1]{\mathbb{R}^{#1}{}}

\newcommand{\integers}[1]{\mathbb{Z}^{#1}{}}

\newcommand{\positivesd}[1]{\mathbb{S}_+^{#1}{}}

\newcommand{\bmat}[1]{\begin{bmatrix}#1\end{bmatrix}}

\newcommand{\transpose}{^{\top}}

\DeclarePairedDelimiter{\norm}{\lVert}{\rVert}

\newcommand{\vct}[1]{\mathbf{#1}}

\DeclareMathOperator{\rank}{rank}

\DeclareMathOperator*{\argmax}{\arg\!\max}

\DeclareMathOperator{\trace}{tr}

\DeclareMathOperator{\vecop}{vec}
\DeclareMathOperator{\stack}{stack}

\newcommand{\subjectto}{\textrm{subject to}\;}

\providecommand{\va}{\vct{a}}

\providecommand{\vb}{\vct{b}}

\providecommand{\vf}{\vct{f}}

\providecommand{\vp}{\vct{p}}

\providecommand{\vq}{\vct{q}}

\providecommand{\vt}{\vct{t}}

\providecommand{\vv}{\vct{v}}

\providecommand{\vx}{\vct{x}}

\providecommand{\vy}{\vct{y}}

\providecommand{\mA}{\vct{A}}
\providecommand{\mB}{\vct{B}}
\providecommand{\mC}{\vct{C}}

\providecommand{\mI}{\vct{I}}

\providecommand{\mK}{\vct{K}}
\providecommand{\mL}{\vct{L}}
\providecommand{\mM}{\vct{M}}

\providecommand{\mQ}{\vct{Q}}
\providecommand{\mR}{\vct{R}}

\providecommand{\mT}{\vct{T}}

\providecommand{\mX}{\vct{X}}
\providecommand{\mY}{\vct{Y}}
\providecommand{\mZ}{\vct{Z}}

\providecommand{\cC}{\mathcal{C}}

\providecommand{\cR}{\mathcal{R}}
\providecommand{\cS}{\mathcal{S}}

\providecommand{\cV}{\mathcal{V}}

\providecommand{\cX}{\mathcal{X}}
\providecommand{\cY}{\mathcal{Y}}

\usepackage{units}

  \newcommand{\newcolorlabel}[2]{%
  \expandafter\newcommand\csname #1\endcsname[1]{%
    \tikz[baseline]{\node[text=white,fill=#2,anchor=base,text height=1.3ex,text depth=0.1ex,font=\sffamily\bfseries]{##1}}}%
}

\newcommand{\newcommenter}[2]{%
  \expandafter\newcommand\csname #1\endcsname[1]{%
    \fcolorbox{#2}{#2}{\color{white}\textsf{\textbf{#1}}}
    {\color{#2}##1}}%
  \expandafter\newcommand\csname at#1\endcsname{%
    \fcolorbox{#2}{#2}{\color{white}\textsf{\textbf{@#1}}}
    {\color{#2}}}%
  \expandafter\newcommand\csname #1cite\endcsname[1]{%
    \csname #1\endcsname {[##1]}
  }%
  \expandafter\newcommand\csname #1ref\endcsname[1]{%
    \csname #1\endcsname {$\blacktriangleright$##1}
  }%
  \expandafter\newcommand\csname #1hl\endcsname[2]{%
    \colorbox{#2}{\color{white}\textsf{\textbf{#1}}}\sethlcolor{Azure2}\hl{##2}~%
    \expandafter\ifx\csname commentarrow\endcsname\relax$\leftarrow$\else \commentarrow[#2]\fi~%
    {\color{#2}##1}}%
  \expandafter\newcommand\csname #1st\endcsname[2]{%
    \colorbox{#2}{\color{white}\textsf{\textbf{#1}}}\sout{##2}~%
    \expandafter\ifx\csname commentarrow\endcsname\relax$\leftarrow$\else \commentarrow[#2]\fi~%
    {\color{#2}##1}}%
}
\newcommenter{TODO}{DodgerBlue1}
\newcommenter{rtron}{OliveDrab2}

\usepackage{comment}

\usepackage{pdfcomment}

\usepackage{soul}

\usepackage[normalem]{ulem}

\usepackage{csquotes}

\usepackage{suffix}

\usepackage{environ}

\makeatletter
\newsavebox{\boxifnotempty}
\newcommand{\displayifnotempty}[3]{\sbox\boxifnotempty{#2}\setbox0=\hbox{\usebox{\boxifnotempty}\unskip}%
  \ifdim\wd0=0pt
  \else
  #1\usebox{\boxifnotempty}#3%
  \fi%
}

\newcommand{\ifempty}[2]{\setbox0=\hbox{#1\unskip}%
  \ifdim\wd0=0pt%
  #2%
  \fi%
}

\newcommand{\ifnotempty}[2]{\setbox0=\hbox{#1\unskip}%
  \ifdim\wd0>0pt%
  #2%
  \fi%
}
\makeatother

\newcommand{\switchifempty}[3]{\sbox\boxifnotempty{#1}\setbox0=\hbox{\usebox{\boxifnotempty}\unskip}%
  \ifdim\wd0=0pt{}%
  #2%
  \else{}%
  #3%
  \usebox{\boxifnotempty}%
  \fi%
}

\makeatletter
\@ifundefined{chapter}{\usepackage{algorithm}}{\usepackage[chapter]{algorithm}}
\makeatother
\usepackage{algorithmicx}
\usepackage{algpseudocode}
\makeatother%

\usepackage{scrlfile}

\makeatletter
\newcommand*\newstoreddef[1]{
  \BeforeClosingMainAux{%
    \immediate\write\@auxout{%
      \string\restoredef{#1}{\csname #1\endcsname}%
    }%
  }%
}
\newcommand*{\restoredef}[2]{
  \expandafter\gdef\csname stored@#1\endcsname{#2}%
}
\newcommand*{\storeddef}[1]{
  \@ifundefined{stored@#1}{0}{\csname stored@#1\endcsname}%
}
\makeatother

\usepackage{pageslts}
\NewEnviron{tee}{\BODY\typeout{Marker Tee [start] ^^J \BODY ^^JMaker Tee [end]}}

\usepackage{cleveref}
\usepackage{nameref}

\usepackage[displaymath, mathlines,switch]{lineno}
\newcommand{\tauset}{\text{\boldmath{$\tau$}}}
\newcommenter{todo}{Firebrick1}
\newcommenter{tomwu}{blue}
\newcommand{\ip}[2]{\left\langle #1,\, #2 \right\rangle}
\newcommand{\robotpart}[1]{\tikz[baseline=-3pt]{\node[draw,circle,minimum size=4mm,font=\sffamily\footnotesize,inner sep=0pt]{#1};}}
\usepackage{environ}    

\newif\ifshowfullcontents
\showfullcontentstrue

\newcommand{\ProofPointer}{%
\begin{proof}
  Please refer to \cite{wu2025arxivfull}
  \end{proof}
}

\NewEnviron{pf}{%
  \ifshowfullcontents
    \begin{proof}\BODY\end{proof}%
  \else
     \ProofPointer\par\medskip
  \fi
}

\title{\LARGE \bf
  Certifiably Optimal Estimation and  Calibration in Robotics\\via Trace-Constrained Semi-Definite Programming
}

\author{Liangting Wu and Roberto Tron
  \thanks{The authors are with the Department of Mechanical Engineering,
    Boston University, 110 Cummington Mall, MA 02215, United States
    {\tt\small tomwu@bu.edu}, {\tt\small tron@bu.edu}.
    The authors gratefully acknowledge the support by NSF award FRR-2212051.}%
}

\begin{document}

\maketitle
\thispagestyle{empty}
\pagestyle{empty}

\begin{abstract}
  Many nonconvex problems in robotics can be relaxed into convex formulations via Semi-Definite Programming (SDP) that can be solved to global optimality. The practical quality of these solutions, however, critically depends on rounding them to rank-1 matrices, a condition that can be challenging to achieve. In this work, we focus on trace-constrained SDPs (TCSDPs), where the decision variables are Positive Semi-Definite (PSD) matrices with fixed trace values. We show that the latter can be used to design a gradient-based refinement procedure that projects relaxed SDP solutions toward rank-1, low-cost candidates. 
  We also provide fixed-trace SDP relaxations for common robotic quantities such as rotations, translations, and a modular \emph{virtual robot} abstraction that simplifies modeling across different problem settings. 
  We demonstrate that our trace-constrained SDP framework can be applied to many robotics tasks, and we showcase its effectiveness through simulations in Perspective-$n$-Point (PnP) estimation, hand-eye calibration, and dual-robot system calibration.
\end{abstract}

\section{Introduction}

\textbf{\textit{SDP relaxation.}} Many problems in robotics can be encoded as a Polynomial Optimization Problem (POP), where the objective function and constraints are polynomials of the variables.
It is well known that such problems are generally non-convex and can be hard to solve optimally. A common approach is via Semi-Definite Relaxations (SDRs), where a relaxed convex problem is solved. This is typically done by constructing a lifted vector $\vx$ using the variables and a moment matrix $\mM=\vx\vx\transpose$ such that the objective function and constraints in the POP are linear functions of $\mM$. SDRs have been popular because: 
\begin{enumerate}[label=(\roman*)]
    \item Relaxations are convex, and can generally be solved with many off-the-shelf solvers. 
    \item The Lagrangian dual of the SDP can provide a certificate of global optimality \cite{yang2020teaser,yang2022certifiably}.
\end{enumerate}

\textbf{\textit{Tightness of SDR.}}
In general, a solution $\vx$ to the original POP can be only extracted when $\mM=\vx\vx\transpose$, i.e., when $\rank(\mM)=1$ \cite{luo2010semidefinite}; however, this condition is typically dropped in the SDR to make the relaxation convex. There exist special cases, such as when the number of constraints is less than $2$, for which solutions to the SDR are also rank-1 \cite{shapiro1982rank,barvinok1995problems,pataki1998rank}; in this case the SDR is said to be \emph{tight}. Many works investigate techniques for making SDR tight by adding \emph{redundant constraints} to the relaxed SDP \cite{yang2022certifiably,dumbgen2024toward}; these constraints are redundant in the original POP, but enforce internal structures on the moment matrix that are lost during the relaxation. \emph{Lassere's hierarchy} \cite{lasserre2001global} provides a way to systematically make a POP SDR tight with a finite (but possibly quite large) number of additional variables and constraints \cite{nie2014optimality}. In \cite{dumbgen2024toward}, the redundant constraints are found automatically by computing the nullspace of a data matrix obtained from using feasible samples.

\textbf{\textit{Scalability.}}
In practice, SDP solvers do not scale favorably with the size of the moment matrix $\mM$. It is therefore important to use sparsity to reduce the size of the problem when many of the entries in $\mM$ do not appear in the original POP \cite{waki2006sums,wang2021chordal}.

\textbf{\textit{Trace-constrained SDP.}} There exists some work exploring the role of constant-trace constraints \cite{helmberg2000spectral,yurtsever2021scalable} for tightness and for  obtaining rank-1 solutions. With constant trace, the SDP with rank-1 constraint can be cast as an eigenvalue optimization problem, enabling the use of gradients of eigenvalues to project solutions onto the rank-1 set. In \protect\cite{mai2022exploiting}, it is shown that any POP with a ball constraint can be relaxed to an SDP with a constant trace constraint. 

The present paper generalizes and provides a more systematic approach to our previous work on Trace-Constrained SDP relaxations for problems in Inverse Kinematics \cite{wu2023cdc,wu2024tac_archive}, and Visual Inverse Kinematics \cite{wu2025visual}.
In particular, we extend the framework to estimation and calibration problems that were not considered in those works. We additionally provide a larger library of relaxations for common quantities in robotics, in some instances (e.g., distance-and-direction constraints), using smaller SDP embedding. Compared to previous work on TCSDP (e.g., \cite{mai2022exploiting, mai2023hierarchy}), we provide specific formulations tailored to robotics problems.


\textit{\textbf{Paper contributions.}} Our key contributions include:
\vspace{-10pt}
\begin{itemize}
    \item We introduce constant-trace semidefinite variables for relaxations of rotation matrices $\mathrm{SO}(3)$ and translations $\mathbb{R}\times\mathcal{S}^2$, which are tightened using structural constraints in addition to the trace constraint, and show that the original manifold variables can be exactly recovered when these constant-trace variables are rank-1.
    \item We provide additional techniques based on ideas from Control Barrier Functions (CBF) to accelerate the solver of \cite{wu2024tac_archive}, which, unlike the original method, can recover solutions that are not only low-rank but also low-cost.
    \item We introduce the SP robot, a virtual kinematic module that models diverse robotics problems, and show that solving its kinematics yields optimal solutions for PnP estimation, hand-eye calibration, and dual-robot calibration problems.
\end{itemize}

\section{Preliminaries}
\subsection{Notation}
Vectors and matrices are denoted by bold-face lowercase and uppercase letters. The reference frames are denoted by sans-serif letters. The operator $\trace(\cdot)$ denotes the trace. A PSD matrix is denoted by $\succeq0$. A list of notations is in Rable \ref{tab:notations}.
\begin{table}[h]
    \centering
    \begin{tabular}{ll}
        $\mathsf{w}$ & World reference frame\\
        ${}^{\mathsf{b}_1}{\mR}_{\mathsf{b}_2}$ & Rotation matrix from reference frame $\mathsf{b}_2$ to $\mathsf{b}_1$ \\
        ${}^{\mathsf{b}_1}{\vt}_{\mathsf{b}_2}$ & Translation from reference frame $\mathsf{b}_2$ to $\mathsf{b}_1$\\
        ${}^{\mathsf{b}_1}{\mT}_{\mathsf{b}_2}$ & Rigid transformation from reference frame $\mathsf{b}_2$ to $\mathsf{b}_1$ \\
        $\mR_{\mathsf{b}_2}$ & Rotation from frame $\mathsf{b}_2$ to the world frame\\
        $\mR^{(i)}$ & The $i$-th column of the matrix $\mR$\\
        $\mK$ & Camera intrinsic matrix\\
        $\mM(a:b,c:d)$ &{\scriptsize Block in matrix $\mM$ for rows $a$ to $b$ and columns $c$ to $d$}\\
        $\{\vv_i\}_n$ & A set of members consisting $\vv_1,\dots,\vv_n$\\
        $\stack\bigl(\{\vv_i\}_n\bigr)$ & {\scriptsize The vertical concatenation of all the members within $\{\vv_i\}_n$}\\
        $\vecop(\mM)$ & {\scriptsize Vectorization, $\stack(\{\vecop(\mM_i)\}_n)$ when $\mM=\{\mM_i\}_n$}\\
        $\lambda_1(\mM)$ & \makecell[l]{\scriptsize Largest eigenvalue of $\mM$,\\ $\sum_i^n(\{\lambda_1(\mM_i)\}_n)$ when $\mM=\{\mM_i\}_n$}\\
        $\positivesd{d}$ &Space of $d\times d$ positive semi-definite matrices\\
        $\mathrm{SO}(d)$ & Special orthogonal group in dimension $d$\\
        $\mathcal{S}^n$ & The $n$-sphere: $\mathcal{S}=\{\vx\in\real{n+1}|\norm{\vx}_2^2=1\}$\\
        $f^\star$ &An optimal cost of function $f$
    \end{tabular}
    \caption{Table of notations}
    \label{tab:notations}
\end{table}

\vspace{-20pt}
\subsection{SDP Relaxations}
Define $\mY=\{\mY_i|\mY_i\in\positivesd{d_y,i}\}_{n_y}$ as a collection of matrices. Denote $N=\dim(\vecop(\mY))$.
Given $\mathbf{Q}\succeq \mathbf{0}$, $\mathbf{c},\va_j\in\real{N}$, consider the following quadratic SDP:
\begin{problem}[Trace-Constrained Relaxed SDP]\label{prob:sdp-relax}
{\footnotesize
    \begin{subequations}\label{eq:sdp-relax}
        \begin{align}
            \min_{\mY=\{\mY_i\}_{n_y}}\quad &f(\mY):=\vy\transpose\mathbf{Q}\,\vy \;+\; \mathbf{c}\transpose\vy\\ 
            \text{s.t.}\quad & \va_j\transpose\vy=b_j, &j=1,\dots,n_e,\label{eq:relax sdp linear}\\
        & \trace(\mY_i) = \bar{\lambda}, &i\in\{1,\dots,n_y\},\label{eq:relax sdp fix trace}\\
        & \mY_i\in\positivesd{d_{y,i}}&i\in\{1,\dots,n_y\},\label{eq:relax sdp psd}
        \end{align}
    \end{subequations}
    }
\end{problem}
where $\{d_{y,i}\}$, $n_y$, and $n_e$ are the dimensions of the embedding, the number of variables, and the number of constraints. The matrix $\mY_i$ is positive semi-definite with a constant trace $\bar{\lambda}$.

Unlike standard SDP formulations, the constant-trace constraint \eqref{eq:relax sdp fix trace} bounds the largest eigenvalue $\lambda_1$ of $\mY_i$, linking rank-1 conditions to maximizing $\lambda_1$ (see \Cref{sec:gradient-based-rank-min}).


We now convert Problem \ref{prob:sdp-relax} into standard form.
Define $\mathcal{A}(\mY)=\vb$ as the linear operator of all linear (including the trace) constraints in Problem \ref{prob:sdp-relax} and its adjoint $\mathcal{A}^*:\ip{\mathcal{A}^*(\boldsymbol{\rho})}{\mathbf{Y}} \coloneqq \sum_{i=1}^n \ip{\big(\mathcal{A}^*(\boldsymbol{\rho})\big)_i}{\mathbf{Y}_i}$. Decompose $\mQ=\mL\transpose\mL$, define the expression $\vy \coloneqq \vecop(\mathbf{Y})$, and write the following standard form SDP equivalent to Problem \ref{prob:sdp-relax}.
\begin{problem}[Primal, standard form SDP]\label{prob:standard sdp}
    \begin{subequations}
    \begin{align}
\min_{\{\mathbf{Y}_i\succeq \mathbf{0}\},\, t} \quad
& t \;+\; \mathbf{c}\transpose\vy \label{eq:sdp_primal_epi_obj}\\[0.5ex]
\text{s.t.}\quad
& \mathcal{A}(\mathbf{Y}) \;=\; \mathbf{b}, \qquad \mathbf{Y}_i \succeq \mathbf{0}\ \ \forall i, \label{eq:sdp_primal_epi_constr1}\\[0.5ex]
& \begin{bmatrix}
t & (\mathbf{L}\vy)\transpose \\
\mathbf{L}\vy & \mathbf{I}
\end{bmatrix}\succeq \mathbf{0}. \label{eq:sdp_primal_epi_constr2}
\end{align}
\end{subequations}
\end{problem}

The epigraph variable $t\in\real{}$ and \eqref{eq:sdp_primal_epi_constr2} imply $t \ge \vy\transpose\mathbf{Q}\,\vy$. Define $\cV^*:\ip{\mathcal{V}^*(\mathbf{r})}{\mathbf{Y}} = \mathbf{r}\transpose\vecop(\mathbf{Y})$. The dual problem of Problem \ref{prob:standard sdp} is
\begin{problem}[Dual SDP]\label{prob:dual}
{\footnotesize
    \begin{subequations}
    \begin{align}
\max_{\boldsymbol{\rho},\,\mathbf{S}=\{\mathbf{S}_i\},\,\mathbf{Z}} \quad
& d(\boldsymbol{\rho},\mathbf{Z}):=\boldsymbol{\rho}\transpose\mathbf{b} \;-\; \operatorname{tr}(\mathbf{Z})\;+\;1 \label{eq:dual_obj}\\[0.25ex]
\text{s.t.}\quad
& \mathbf{Z}=\begin{bmatrix}1 & \mathbf{z}\transpose\\ \mathbf{z} & \mathbf{Z}_{22}\end{bmatrix}, \label{eq:dual_Z}\\[0.5ex]
& \mathcal{V}^*\!\big(\mathbf{c}-2\mathbf{L}\transpose\mathbf{z}\big)\;=\;\mathcal{A}^*(\boldsymbol{\rho})\;+\;\mathbf{S}, \\
& \mathbf{S}_i\succeq \mathbf{0}\ \ \forall i=1,\dots,n_y, \label{eq:dual_stat_final}
\end{align}
\end{subequations}
}
\end{problem}
where $\boldsymbol{\rho}\in\real{n_e}, \mathbf{Z}, \mathbf{S}$ are dual variables.
\ifshowfullcontents
See the Appendix for the derivation.
\fi

It is well known that $f^{\star}\geq d^{\star}$ holds and $f^\star-d^\star$ is called the duality gap. For a feasible solution $(\hat{\mY},\hat{t})$, we can certify its optimality directly by the Karush--Kuhn--Tucker (KKT) conditions.
In particular, if there exist dual variables
$(\hat{\boldsymbol{\rho}}, \{\hat{\mathbf{S}}_i\}, \hat{\mathbf{Z}})$ such that

{\scriptsize
\begin{subequations}
    \begin{align}
&\ip{\hat{\mathbf{Z}}}{\begin{bmatrix}
\hat{t} & (\mathbf{L}\hat{\vy})^{\top}\\[0.3ex]
\mathbf{L}\hat{\vy} & \mathbf{I}
\end{bmatrix}} = 0,\quad \ip{\hat{\mathbf{S}}_i}{\mathbf{Y}_i^{\star}} = 0,
\quad \forall i=1,\dots,n_y;\label{eq:kkt_slackness}\\[1ex]
& \hat{\mathbf{S}}_i \succeq \mathbf{0}\ \ \forall i,
\qquad \hat{\mathbf{Z}} \succeq \mathbf{0}, \label{eq:kkt_dual_feas}
\end{align}
\end{subequations}
}
where $\hat{\vy}:=\vecop(\hat{\mY})$, then \emph{strong duality} holds and the primal and dual optimal values coincide.
In this case, the duality gap is zero and $(\hat{\mY},\hat{t})$ is a global optimal solution.

\section{Rank and Cost Optimization of TCSDP}\label{sec:rank min}

In this section, we present a rank minimization method to project relaxed SDP solutions onto rank-1 matrices, followed by a cost-reduction method for rank-1 solutions.
\subsection{A gradient-based approach for rank minimization}\label{sec:gradient-based-rank-min}
This subsection discusses the rank minimization strategy from \cite{wu2023cdc} that applies to any fixed-trace matrix.

For a PSD matrix $\mM\in\positivesd{d}$ with fixed trace, $\rank(\mM)=1$ if and only if
\begin{equation}\label{eq:matrix rank}
    \mM = \argmax\left(\lambda_1(\mM)\right),
\end{equation}
where $\lambda_1(\cdot)$ indicates the largest eigenvalue of its matrix argument. Intuitively, since the trace is also the sum of eigenvalues, the sum $\sum_{l=1}^{d}\lambda_l(\mM)$ is also fixed. This implies that when $\lambda_1$ is maximized, we have $\lambda_l(M)=0, \forall l = 2,\dots,d$, hence $M$ is rank-1; see \cite{wu2023cdc} for a proof.

Using \eqref{eq:matrix rank}, we can project a matrix $\mM$ to the rank-1 set by increasing $\lambda_1(\mM)$. Fortunately, the function $\lambda_1(\mM)$ is differentiable on $\mM$ (so long as $\lambda_1(\mM)$ has multiplicity 1) and the gradient $\nabla\lambda_1(\mM)$ can be computed \cite[Theorem 1]{magnus1985}.
This enables the projected gradient-descent approach of \cite{wu2023cdc} which increases $\lambda_1(\mM)$ by solving the following SDP at every iteration:
\begin{problem}[Rank Minimization Update]\label{prob:simple rank min update}
    \begin{subequations}\label{eq:simple rank min update}
    {\footnotesize
        \begin{align}
            \min_{\Delta\mY^{k},c}\quad &f(\mY^{k-1}+\Delta\mY^k)+\gamma_c c\label{eq:simple rank min obj}\\
        \text{s.t.}\quad 
        & \vecop(\Delta\mY^{k})\transpose\nabla\lambda_{1}(\mY^{k-1}) {\geq} (c{-}1)(\lambda_{1}(\mY^{k-1}){-}\bar{\lambda}_s)\label{eq:simple rm rankconstraint Y}\\
        & c\in [0,1]\label{eq:simple rm c constraint}\\
        & \mY_i^{k-1}+\Delta\mY_i^k \in \cY, \forall i=1,\dots,n_y\label{eq:simple rm relaxed convex set}
        \end{align}
        }
    \end{subequations}
\end{problem}
The scalar $\gamma_c$ is a weight factor, and $\bar{\lambda}_s$ is the sum of the fixed trace value for all $\mY_i$. The constraint $\cY$ is the feasible set \eqref{eq:relax sdp linear}-\eqref{eq:relax sdp psd}. The solution to \eqref{eq:simple rank min update} computes an increment $\Delta \mY^{k}$ such that  $\mY^k=\Delta \mY^{k} +\mY^{k-1}$. The constraints \eqref{eq:simple rm rankconstraint Y} and \eqref{eq:simple rm c constraint} imply that 
the concave function $\bar{\lambda}_s-\lambda_1(\mY^k)$ decreases at a geometric rate \cite[Sec. VII.D]{wu2024tac_archive}.
Problem \ref{prob:simple rank min update} tries to find the update that minimizes the cost subject to the constraints, but the latter take precedence; as a result, it might not converge to a constrained minimizer of $f$.

\subsection{Low-rank channel}\label{sec:low-rank-channel}
As shown in \cite{wu2024tac_archive,wu2025visual}, solving a sequence of Problem \ref{prob:simple rank min update} often yields solutions satisfying $\bar{\lambda}_s-\sum_i^{n_y}\lambda_1(\mY_i)\approx0$, i.e., approximately rank-1 under a given tolerance. However, such rank-1 solutions are not necessarily optimal. In practice, the cost $f(\mY)$ typically increases during rank minimization \eqref{eq:simple rank min update}, with the rate of increase influenced by the weight $\gamma_c$. Moreover, applying Problem \ref{prob:simple rank min update} to reduce cost after reaching the rank-1 set leads to only minor improvements, since the updates effectively attempt linear movements along a curved boundary defined by \eqref{eq:simple rm rankconstraint Y}, without leaving the boundary itself.

To account for the above limitations, we provide a novel iterative approach to find a low-cost solution while maintaining the rank-1 property.

We note two observations. \begin{enumerate*}[label=(\roman*)]
    \item It is well known that the optimal solution $\mY^{\star}$ is also rank-1.
    \item The rank-1 set is on the boundary of the feasible set \eqref{eq:simple rm relaxed convex set} (this is because the boundary of the PSD cone is the singular matrices).
    \end{enumerate*}

Our strategy is to move $\mY$ along the rank-1 set while minimizing the cost towards $\mY^{\star}$ by introducing a parameter $\gamma$ in Problem \ref{prob:simple rank min update}.
\vspace{1pt}
\begin{problem}(Low-rank channel update)\label{prob:low_rank_channel}
{\footnotesize
    \begin{subequations}
        \begin{align}
            \min_{\Delta\mY^{k},c}\quad &f(\mY^{k-1}+\Delta\mY^{k})\label{eq:obj_low_rank_channel}\\
        \text{s.t.}\quad
        &\vecop(\Delta\mY^{k})\transpose\nabla\lambda_{1}(\mY^{k-1})\geq(c-1)(\lambda_1(\mY^{k-1})-\gamma\bar{\lambda}_s)\label{eq:rank-lower-bound}\\
        & \text{\eqref{eq:simple rm c constraint} and \eqref{eq:simple rm relaxed convex set}} \label{eq:lrc-manifold_low_rank_channel}
        \end{align}
    \end{subequations}
    }
\end{problem}

Problem \ref{prob:low_rank_channel} aims to find a solution such that the updated matrices are \emph{close} to rank-1, summarized in the following proposition.
\begin{proposition}\label{prop:low-rank-channel}
    Given $\gamma\in(0,1)$ and a $\mY^{k-1}$ that satisfies $\lambda_1(\mY^{k-1})\geq \gamma\bar{\lambda}_s$, for an optimal solution $\Delta\mY^k$ of Problem \ref{prob:low_rank_channel}, we have
    \begin{equation}\label{eq:channel}
        \gamma\bar{\lambda}_s\leq\lambda_1(\mY^{k-1}+\Delta\mY^k)\leq\bar{\lambda}_s.
    \end{equation}
\end{proposition}

\begin{pf}
    We start with the first inequality of \eqref{eq:channel}. Inspired by concepts in Control-Barrier-Function (CBF), we define
\begin{equation}\label{eq:cbf_lambda1}
    h(\mY):=\lambda_1(\mY)-\gamma\bar{\lambda}_s
\end{equation}
For a sequence $\{\mY^k\}$ and $h(\mY^0)\geq 0$, we have $h(\mY^k)\geq0, \forall k\in\integers{}^+$ if the following condition holds \cite{agrawal2017discrete}[Prop. 4].
\begin{equation}\label{eq:cbf_lambda1_condition}
    \Delta h(\mY^{k})+b h(\mY^{k-1})\geq 0, \forall k\in\integers{}^+, b\in(0,1].
\end{equation}
Substitute \eqref{eq:cbf_lambda1} into \eqref{eq:cbf_lambda1_condition} for $\mY^{k-1}$ and $\mY^{k}$ to get
\begin{equation}\label{eq:cbf_lambda1_Yk_1}
    \begin{aligned}
        \lambda_1(\mY^{k})-\lambda_1(\mY^{k-1})+b(\lambda_1(\mY^{k-1})-\gamma\bar{\lambda}_s)\geq 0
    \end{aligned}
\end{equation}
Because $\lambda_1(\mY)$ is a convex function, we have
\begin{equation}\label{eq:lambda_1_yk_convexity}
    \lambda_1(\mY^{k})\geq \lambda_1(\mY^{k-1})+\vecop(\Delta\mY^{k})\transpose\nabla\lambda_{1}(\mY^{k-1})
\end{equation}
Using \eqref{eq:lambda_1_yk_convexity}, $\lambda_1(\mY^{k-1})>0,\lambda_1(\mY^{k})>0$, and $b h(\mY^{k-1})\geq 0$, we have
{\footnotesize
\begin{equation}\label{eq:cbf_inner_bound_0}
    \begin{aligned}
        \vecop(\Delta\mY^{k})\transpose\nabla\lambda_{1}(\mY^{k-1})+b(\lambda_1(\mY^{k-1})-\gamma\bar{\lambda}_s)\geq 0 \Rightarrow \eqref{eq:cbf_lambda1_Yk_1}.
    \end{aligned}
\end{equation}
}
Replace $b$ with $c\in[0,1)$, the l.h.s. of \eqref{eq:cbf_inner_bound_0} is exactly \eqref{eq:rank-lower-bound}. The implication holds when $c=1$. Thus $h(\mY^k)\geq0$ and the left inequality holds when \eqref{eq:rank-lower-bound} and \eqref{eq:lrc-manifold_low_rank_channel} hold for $\mY^{k-1}$.

The second inequality of \eqref{eq:channel} is enforced by the constant-trace and the PSD constraints.
\end{pf}

Proposition \ref{prop:low-rank-channel} establishes that the sum of the largest eigenvalues of the updated solution lies between the bounds in \eqref{eq:rank-lower-bound} and \eqref{eq:lrc-manifold_low_rank_channel}, thereby defining a \emph{low-rank channel} of width determined by $\gamma$ within which the solution can be steered toward lower cost.

\textit{\textbf{Tolerance Scheduling.}}
    Iteratively solving Problem \ref{prob:low_rank_channel} improves the cost, but progress can be slow when the rank-minimized solution is far from low-cost; to address this, we propose replacing constraint \eqref{eq:rank-lower-bound} with \eqref{eq:scheduling rank constraint} to more efficiently reduce the cost of a rank-1 solution.
    
    {\footnotesize
    \begin{equation}\label{eq:scheduling rank constraint}
            \vecop(\Delta\mY^{k})\transpose\nabla\lambda_{1}(\mY^{k-1}) {\geq} (c{-}1)(\lambda_{1}(\mY^{k-1}){-}\bar{\lambda}_s)-\sigma^k.
    \end{equation}
    }
    
For $\bar{\sigma}>0$, we define a decreasing tolerance sequence ${\sigma^k}$ for the rank reduction constraint \eqref{eq:simple rm rankconstraint Y} that reaches $\bar{\sigma}$ in finite steps. Unlike Problem \ref{prob:low_rank_channel}, which strictly bounds the rank, tolerance scheduling uses the soft constraint \eqref{eq:scheduling rank constraint} to gradually reduce rank while allowing cost-reducing updates of larger magnitude. In practice, scheduling can be combined with the low-rank channel: first moving a rank-1 solution to a low-cost region, then fine-tuning with the channel. An example ${\sigma^k}$ and simulations are presented in Section \ref{sec:sim}.

\section{Trace-Constrained SDP Relaxations}\label{sec:sdp relax}
In this section, we show how fixed-trace PSD matrices can represent both rotation matrices and rigid translations.

\subsection{Relaxation of $\mathrm{SO}(3)$}\label{sec:so3 relax}
We start with a brief review of a relaxation of $\mathrm{SO}(3)$ originally introduced in \cite{wu2023cdc}.

Consider a problem with $n_y$ rotation matrices as variables. For each rotation matrix $\mR_i\in\mathrm{SO}(3)$, we define a decision variable $\mY_i$.
\begin{definition}
A matrix $\mY_i$ can be built from $\mR_i$ as
    \begin{equation}\label{eq:Ystructure}
  \begin{aligned}
    \mY_{i} = \bmat{\mR_{i}^{(1)}\\\mR_{i}^{(2)}\\1}\bmat{\mR_{i}^{(1)}\\\mR_{i}^{(2)}\\1}\transpose\in\real{7\times7}.
  \end{aligned}
\end{equation}
 We define the constraint $\mY_i\in\cY$ as
\begin{subequations}
    \begin{align}
        \trace(\mY_i(1:3,1:3))=\trace(\mY_i(4:6,4:6))&=1\label{eq:trace constraint 1}\\
        \trace(\mY_i(1:3,4:6))&=0\label{eq:trace constraint 2}\\
        \mY_i(7,7)&=1\label{eq:trace constraint 3}\\
        \mY_i\succeq 0\label{eq:Y PSD constraint}
    \end{align}
\end{subequations}
\end{definition}

The matrix $\mY_i$ contains all elements of $\mR_i^{(1),(2)}$, $\mR_i^{(1)}\cdot \mR_i^{(2)}$, which are used in \eqref{eq:trace constraint 1}-\eqref{eq:trace constraint 3} to enforce $\mR_i^{(1)}\cdot \mR_i^{(2)}=0$ and $\norm{\mR_i^{(1),(2)}}=1$ --- equations required for $\mR_i\in\mathrm{SO}(3)$. Observe that \eqref{eq:trace constraint 1}-\eqref{eq:trace constraint 3} also fix the trace of $\mY_i$. The matrix $\mY_i$ contains also all the elements of the third column through $\mR_i^{(3)}=\mR_i^{(1)}{}\times\mR_i^{(2)}$.

We can represent all three columns of $\mR_i$ using a linear mapping $g(\cdot):=\real{7\times7}\rightarrow\real{3\times3}$, $\mR_i=g(\mY_i)$; this map $g(\cdot)$ does not guarantee the rotation constraint $\mR_i\in\mathrm{SO}(3)$ unless accompanied by additional constraints:
\begin{theorem}\label{thm:rot manifold}
    A matrix $\mR_i=g(\mY_i)$ is a rotation matrix (i.e., $\mR_i\in\mathrm{SO}(3)$) if and only if $\mY_i\in\cY$ and $\rank(\mY_i)=1$.
\end{theorem}
See \cite{wu2024tac_archive} for a proof.

Theorem \ref{thm:rot manifold} shows that our relaxation of $\mathrm{SO}(3)$ is tight on the rank-1 set, meaning that we can solve for a solution on the relaxed set and exactly recover the corresponding rotations.

\subsection{Relaxation of a subset of $\real{}\times\mathcal{S}^2$}\label{sec:translation relax}
Consider a space of unit vectors $\vv\in\cS^2$ and a set of real numbers $\tau\in[0,1]$. The tuple $(\tau,\vv)$ forms a subset of $\real{}\times\mathcal{S}^2$ (in many applications, this can be the distance and direction of a 3-D vector).
We now introduce a way to formulate variables in this subset as fixed-trace matrices, similar to what is done for rotations in Section \ref{sec:so3 relax}.

\begin{definition}\label{def:prismatic relaxation}
Given $\tau\in [0,1]$ and $\vv=\bmat{v_1 &v_2 &v_3}\transpose\in\real{3}$, define three matrices $\{\mY_{\tau,l}\}_3$ as:
\begin{equation}
\mY_{\tau,l}=
    \bmat{\sqrt{\tau}v_l\\\sqrt{1-\tau}v_l\\\sqrt{\tau}\\\sqrt{1-\tau}}\bmat{\sqrt{\tau}v_l\\\sqrt{1-\tau}v_l\\\sqrt{\tau}\\\sqrt{1-\tau}}\transpose, l\in\{1,2,3\},
\end{equation}
and define the constraint $\{\mY_{\tau,l}\}_3\in\cY_{\tau}$ as:
\begin{enumerate}
{\footnotesize
    \item $\mY_{\tau,l}\succeq 0$;\label{itm:ytau psd}
    \item $\sum_{l}^3 \trace(\mY_{\tau,l})=4$;\label{itm:ytau trace all}
    \item $\mY_{\tau,l}(3,3)+\mY_{\tau,l}(4,4)=1$;\label{itm:ytau trace 34}
    \item $\mY_{\tau,l}(1,4)=\mY_{\tau,l}(2,3)$;\label{itm:ytau offdiag eq1}
    \item $\mY_{\tau,l}(1,3),\mY_{\tau,l}(1,4),\mY_{\tau,l}(2,3),\mY_{\tau,l}(2,4)\in[-1,1]$;\label{itm:ytau element bound}
    \item $\mY_{\tau,l}(3,3)\in [0,1],\quad \mY_{\tau,l}(3,4)\in[0,1]$;\label{itm:ytau tau bound}
    \item $\mY_{\tau,l}(3,3)=\mY_{\tau,l'}(3,3)$, \quad $\mY_{\tau,l}(3,4)=\mY_{\tau,l'}(3,4)$;\label{itm:ytau cross mat tau eq}
    \item $\sum_{l=1}^3\mY_{\tau,l}(1,2)=\mY_{\tau,1}(3,4)$;\label{itm:ytau unit vec}
    \item $\sum_{l=1}^3\mY_{\tau,l}(1,1){=}\mY_{\tau,1}(3,3),\  \sum_{l'=1}^3\mY_{\tau,l'}(2,2){=}\mY_{\tau,1}(4,4)$;\label{itm:ytau trace 13 24}
    }
    \end{enumerate}
    unless appearing in a sum, the constraints should hold for all $l,l'\in\{1,2,3\}$, $l\neq l'$.
\end{definition}

Definition \ref{def:prismatic relaxation} parallels \cite{wu2024tac_archive}, where the matrix is the outer product of $\bmat{\sqrt{\tau}\vv&\sqrt{1-\tau}\vv&\sqrt{\tau}&\sqrt{1-\tau}}\transpose$; however, $\{\mY_{\tau,l}\}_3$ is smaller (48 vs. 64 variables) by excluding unused cross-terms while retaining the necessary submatrices, with the role of the constraints $\cY_\tau$ explained below:
\begin{itemize}
    \item Constraint \ref{itm:ytau trace all} fixes the \emph{sum} of traces of $\{\mY_{\tau,l}\}_3$ and \ref{itm:ytau trace 34} fixes the trace of each $2\times2$ diagonal submatrix.
    \item Constraint \ref{itm:ytau offdiag eq1} restricts the off-diagonal equalities of each $\mY_{\tau,l}$ and \ref{itm:ytau element bound} restricts the bounds of these elements.
    \item Constraint \ref{itm:ytau tau bound} restricts the joint position limit $\tau\in[0,1]$.
    \item Constraint \ref{itm:ytau cross mat tau eq} enforces the consistency of $\tau$ and $\sqrt{\tau(1-\tau)}$ between any two $\mY_{\tau,l}$.
    \item Constraints \ref{itm:ytau unit vec} and \ref{itm:ytau trace 13 24} both involve a sum over elements across different $l\in\{1,2,3\}$ based on the fact that $\sum_lv_l^2=1$ when $\norm{\vv}=1$.
\end{itemize}


The following theorem shows that $\tau\in[0,1]$ and $\vv\in\cS^2$ can be recovered linearly from rank-1 $\mY_\tau$.
\begin{theorem}\label{thm:unitlength}
    Define the following function $g_\tau : \real{4\times4\times3}\rightarrow\real{}\times\cS^2$.
    \begin{equation}\label{eq:extract vector}
    \begin{aligned}
        (\tau,\vv) =& g_\tau(\{\mY_{\tau,l}\}_3)\\
        :=&\left(\mY_{\tau,1}(3,3),\stack\{\mY_{\tau,l}(1,3){+}\mY_{\tau,l}(2,4)\}_3\right)
    \end{aligned}
    \end{equation}
    Let $(\tau,\vv) := g_\tau(\{\mY_{\tau,l}\}_3)$.
    When the matrices $\{\mY_{\tau,l}\}_3\in\cY_{\tau}$ and $\rank(\mY_{\tau,l})=1,\forall l$, it holds that
    \begin{equation}\label{eq:unit vector from matrices 2}
        \tau\in[0,1],\text{and}\quad \norm{\vv}=1
    \end{equation}
\end{theorem}
\begin{pf}
    By \eqref{itm:ytau tau bound}, $\tau\in[0,1]$ holds at rank-1 $\mY_\tau^k$.
    When each matrix $\mY_{\tau,l}$ is rank-1, it can be written as
    \begin{equation}\label{eq:rank1 matrix explicit}
        \mY_{\tau,l} {=} \hat{\vy}\hat{\vy}\transpose
    \end{equation}
    where $\hat{\vy}=\bmat{s_1\sqrt{a_l}&s_2\sqrt{b_l}&s_3\sqrt{c_l}&s_4\sqrt{d_l}}\transpose$, and $s_1,$$\dots,$$s_4$ are sign indicator $\pm1$.
   Then the l.h.s. of \eqref{eq:unit vector from matrices 2} equals to
    \begin{equation}\label{eq:lhs terms}
        \sum_{l=1}^3a_lc_l+b_ld_l+2s_1s_2s_3s_4\sqrt{a_l b_l c_l d_l}
    \end{equation}
    Based on constraints \ref{itm:ytau trace 34} and \ref{itm:ytau cross mat tau eq}, we let $\hat{\tau}=c_l, \forall l$ and we have $d_l = 1-\hat{\tau}, \forall l$. By constraint \ref{itm:ytau trace 13 24} we have $\sum_{l=1}^3a_l = \hat{\tau}$ and $\sum_{l=1}^3b_l = 1-\hat{\tau}$. By constraint \ref{itm:ytau unit vec} we have $s_1s_2\sum_{l=1}^3\sqrt{a_lb_l}=s_3s_4\sqrt{c_1d_1}=s_3s_4\hat{\tau}(1-\hat{\tau})$. With these, \eqref{eq:lhs terms} becomes
    \begin{equation}
       \hat{\tau}^2+(1-\hat{\tau})^2+2\hat{\tau}(1-\hat{\tau})= 1
    \end{equation}
\end{pf}

The following proposition shows that when $\{\mY_{\tau,l}\}_3$ is rank-one, the bilinear term $\tau\vv$ can be recovered as a linear function of $\mY_\tau$, ensuring the validity of the product.

\begin{proposition}\label{prop:tauv}
Define the function $g_{\tau v}:\real{4\times4\times3}\rightarrow\real{3}$.
\begin{equation}\label{eq:extract tau vector}
\begin{aligned}
g_{\tau v}(\{\mY_{\tau,l}\}_3)
    :=\stack\{\mY_{\tau,l}(1,3)\}_3
\end{aligned}
\end{equation}
    Let $(\tau,\vv):=g_\tau(\{\mY_{\tau,l}\}_3)$. When the matrices $\{\mY_{\tau,l}\}_3\in\cY_\tau$ and $\rank(\mY_{\tau,l})=1,\forall l$, it holds that
    \begin{equation}\label{eq:tau v and v}
        g_{\tau v}(\{\mY_{\tau,l}\}_3)=\tau\vv 
    \end{equation}
\end{proposition}

\begin{pf}
    We write $\mY_{\tau,l}$ in the form of \eqref{eq:lhs terms} and proving \eqref{eq:tau v and v} is equivalent to prove the following equality,
    \begin{equation}
    s_1s_3\sqrt{a_lc_l}=s_1s_3c_l\sqrt{a_lc_l}+s_2s_4c_l\sqrt{b_ld_l},\quad \forall l=1,\dots,3\label{eq:explicit tau v and v}
    \end{equation}
    By constraint \ref{itm:ytau offdiag eq1} we have $s_2s_3\sqrt{b_lc_l}=s_1s_4\sqrt{a_ld_l}$. Substitute to the r.h.s of \eqref{eq:explicit tau v and v} to get $s_1\sqrt{a_lc_l}(s_3c_l+s_4d_l)$. By constraints \ref{itm:ytau tau bound} and \ref{itm:ytau trace 34} we have $s_3=s_4$ and $c_l+d_l=1$. Substitute to \eqref{eq:explicit tau v and v} and the equality is proved.
\end{pf}

We demonstrate in the following proposition that the rank-1 condition \eqref{eq:matrix rank} can be obtained for every matrix in $\{\mY_{\tau,l}\}_3$ through maximization of a sum of $\lambda_1(\mY_{\tau,l})$.

\begin{proposition}
    When $\sum_i\lambda_1(\mY_{\tau,l})$ is maximized and $\{\mY_{\tau,l}\}_3\in\cY_{\tau}$, i.e.,
    \begin{equation}
        \{\mY_{\tau,l}^{\star}\}_3=\argmax(\sum_l\lambda_1(\mY_{\tau,l}))\in\cY_{\tau},
    \end{equation}
    we have $\rank(\mY^{\star}_{\tau,l})=1, \forall l$.
\end{proposition}

\begin{pf}
    Because the trace is also the sum of the eigenvalues, with the sum of the trace fixed, we have
    \begin{equation}
        \sum_{l=1}^3\lambda_1(\mY_{\tau,l})+\sum_{l=1}^3\sum_{i=2}^4\lambda_i(\mY_{\tau,l})=4.
    \end{equation}
    Since the matrices are PSD, when $\{\mY_{\tau,l}^{\star}\}_3=\argmax(\sum_l\lambda_1(\mY_{\tau,l}))$ we have $\sum_{l=1}^3\sum_{i=2}^4\lambda_i(\mY_{\tau,l}^{\star}) = 0$. Thus $\rank(\mY^{\star}_{\tau,l})=1$, $\forall l\in\{1,2,3\}$.
\end{pf}

\section{Applications}
In this section, we address three robotic problems via virtual robots, introducing the SP robot as a module for building virtual kinematic chains and showing that solutions on these chains correspond to solutions of the original problems.
\begin{figure*}[t]
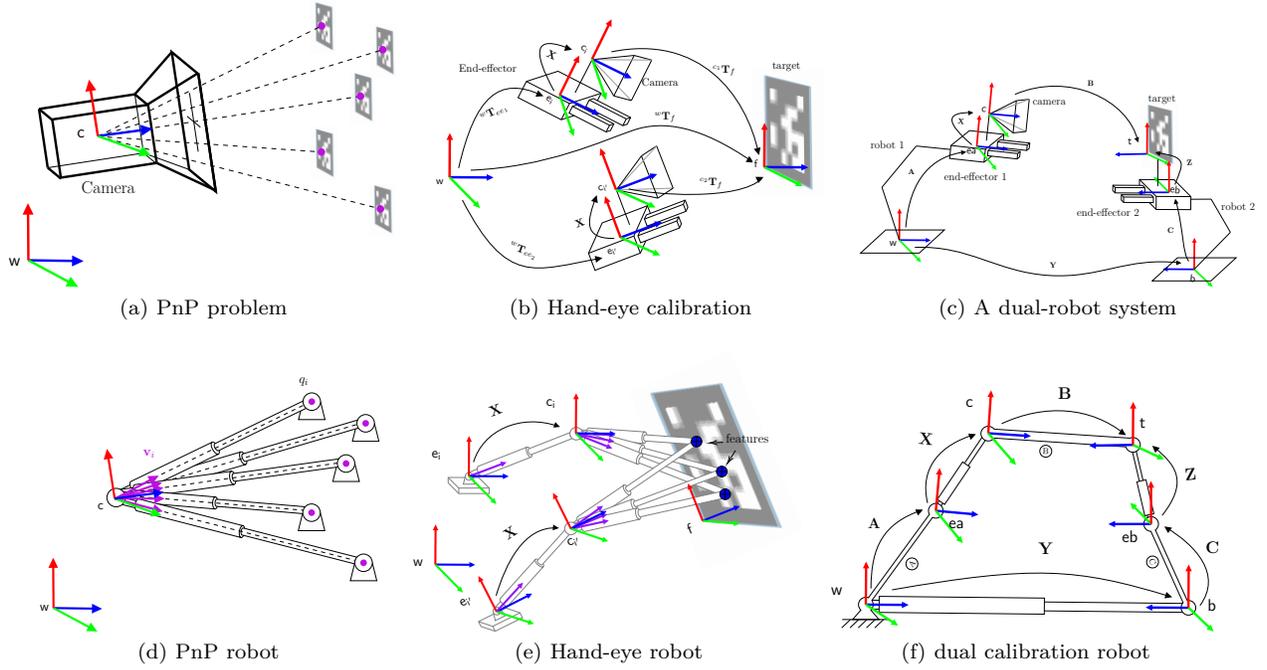

  \centering
  \subfloat[PnP problem]{\resizebox{0.3\textwidth}{!}{\input{pnp_problem_concept}}\label{fig:pnp}}\hspace{6pt}
  \subfloat[Hand-eye calibration]{\resizebox{0.3\textwidth}{!}{\input{handeye}}\label{fig:handeye}}\hspace{6pt}
  \subfloat[A dual-robot system]{\resizebox{0.3\textwidth}{!}{\input{axb_eq_ycz}}\label{fig:axbycz}}

  \subfloat[PnP robot]{\resizebox{0.26\textwidth}{!}{\input{pnp_robot}}\label{fig:pnp-robot}}\hspace{6pt}
  \subfloat[Hand-eye robot]{\resizebox{0.3\textwidth}{!}{\input{hand-eye-kinematic-chain}}\label{fig:hand-eye-kinematic-chain}}\hspace{6pt}
  \subfloat[dual calibration robot]{\resizebox{0.3\textwidth}{!}{\input{dual-hand-eye-robot}}\label{fig:dual-hand-eye-robot}}

  \caption{Some estimation and calibration problems in robotics can be formulated as kinematics problems of virtual robots.}
  \label{problems and virtual robots}
\end{figure*}

\subsection{Rigid translation using forward kinematics of a virtual kinematic chain}
We start with the definition of a robot with two joints.
\begin{definition}
    An \emph{SP robot} has a spherical joint followed by a prismatic joint. The spherical joint centers at the base, and the prismatic joint has an extension limit of $[0,\tau_u]$. The end-effector is the free end of the prismatic joint.
\end{definition}

\begin{figure}[htb]
\vspace{-15pt}
  \centering
  \resizebox{0.38\textwidth}{!}{
  \tikzset{every picture/.style={line width=0.75pt}} 

\begin{tikzpicture}[x=0.75pt,y=0.75pt,yscale=-1,xscale=1]

\draw  [color={rgb, 255:red, 128; green, 128; blue, 128 }  ,draw opacity=1 ][fill={rgb, 255:red, 255; green, 255; blue, 255 }  ,fill opacity=1 ] (245.43,312.62) -- (232.29,285.7) -- (167.21,263.33) -- (164.02,272.62) -- (177.16,299.54) -- (242.24,321.91) -- cycle ; \draw  [color={rgb, 255:red, 128; green, 128; blue, 128 }  ,draw opacity=1 ] (167.21,263.33) -- (180.35,290.25) -- (245.43,312.62) ; \draw  [color={rgb, 255:red, 128; green, 128; blue, 128 }  ,draw opacity=1 ] (180.35,290.25) -- (177.16,299.54) ;
\draw [fill={rgb, 255:red, 255; green, 255; blue, 255 }  ,fill opacity=1 ]   (223.46,267.87) -- (221.21,293.88) -- (185.9,281.74) -- (200.14,259.85) ;

\draw  [fill={rgb, 255:red, 255; green, 255; blue, 255 }  ,fill opacity=1 ] (200.01,262) .. controls (200.01,255.19) and (205.53,249.67) .. (212.34,249.67) .. controls (219.15,249.67) and (224.67,255.19) .. (224.67,262) .. controls (224.67,268.81) and (219.15,274.33) .. (212.34,274.33) .. controls (205.53,274.33) and (200.01,268.81) .. (200.01,262) -- cycle ;
\draw  [fill={rgb, 255:red, 255; green, 255; blue, 255 }  ,fill opacity=1 ] (251.86,265.12) -- (214.47,268.54) .. controls (212.69,268.7) and (210.95,265.56) .. (210.58,261.53) .. controls (210.22,257.5) and (211.36,254.09) .. (213.13,253.93) -- (250.52,250.52) .. controls (252.3,250.35) and (254.03,253.49) .. (254.4,257.53) .. controls (254.77,261.56) and (253.63,264.96) .. (251.86,265.12) .. controls (250.08,265.28) and (248.34,262.15) .. (247.98,258.11) .. controls (247.61,254.08) and (248.75,250.68) .. (250.52,250.52) ;
\draw  [fill={rgb, 255:red, 255; green, 255; blue, 255 }  ,fill opacity=1 ] (290.01,266.82) -- (241.1,271.29) .. controls (238.07,271.57) and (235.11,266.22) .. (234.49,259.34) .. controls (233.86,252.47) and (235.8,246.67) .. (238.83,246.4) -- (287.74,241.93) .. controls (290.76,241.65) and (293.73,247) .. (294.35,253.88) .. controls (294.98,260.75) and (293.04,266.55) .. (290.01,266.82) .. controls (286.99,267.1) and (284.03,261.75) .. (283.4,254.88) .. controls (282.77,248) and (284.71,242.2) .. (287.74,241.93) ;
\draw  [fill={rgb, 255:red, 255; green, 255; blue, 255 }  ,fill opacity=1 ] (393.85,252.15) -- (292.24,261.43) .. controls (290.46,261.59) and (288.73,258.46) .. (288.36,254.42) .. controls (287.99,250.39) and (289.13,246.99) .. (290.9,246.83) -- (392.52,237.54) .. controls (394.29,237.38) and (396.03,240.52) .. (396.4,244.55) .. controls (396.77,248.59) and (395.63,251.99) .. (393.85,252.15) .. controls (392.08,252.31) and (390.34,249.17) .. (389.97,245.14) .. controls (389.6,241.11) and (390.74,237.71) .. (392.52,237.54) ;
\draw [color={rgb, 255:red, 0; green, 0; blue, 0 }  ,draw opacity=1 ] [dash pattern={on 4.5pt off 4.5pt}]  (210.8,260.96) -- (396.4,244.55) ;
\draw  [draw opacity=0][fill={rgb, 255:red, 189; green, 16; blue, 224 }  ,fill opacity=1 ] (391.73,244.55) .. controls (391.73,241.98) and (393.82,239.89) .. (396.4,239.89) .. controls (398.98,239.89) and (401.07,241.98) .. (401.07,244.55) .. controls (401.07,247.13) and (398.98,249.22) .. (396.4,249.22) .. controls (393.82,249.22) and (391.73,247.13) .. (391.73,244.55) -- cycle ;
\draw [color={rgb, 255:red, 0; green, 0; blue, 255 }  ,draw opacity=1 ][line width=2.25]    (212.24,260.89) -- (262.62,278.39) ;
\draw [shift={(267.34,280.03)}, rotate = 199.16] [fill={rgb, 255:red, 0; green, 0; blue, 255 }  ,fill opacity=1 ][line width=0.08]  [draw opacity=0] (8.57,-4.12) -- (0,0) -- (8.57,4.12) -- cycle    ;
\draw [color={rgb, 255:red, 0; green, 255; blue, 0 }  ,draw opacity=1 ][line width=2.25]    (212.24,260.89) -- (235.13,308.15) ;
\draw [shift={(237.31,312.65)}, rotate = 244.16] [fill={rgb, 255:red, 0; green, 255; blue, 0 }  ,fill opacity=1 ][line width=0.08]  [draw opacity=0] (8.57,-4.12) -- (0,0) -- (8.57,4.12) -- cycle    ;
\draw [color={rgb, 255:red, 255; green, 0; blue, 0 }  ,draw opacity=1 ][line width=2.25]    (212.24,260.89) -- (229.86,210.19) ;
\draw [shift={(231.5,205.47)}, rotate = 109.16] [fill={rgb, 255:red, 255; green, 0; blue, 0 }  ,fill opacity=1 ][line width=0.08]  [draw opacity=0] (8.57,-4.12) -- (0,0) -- (8.57,4.12) -- cycle    ;

\draw [color={rgb, 255:red, 144; green, 19; blue, 254 }  ,draw opacity=1 ][line width=2.25]    (211.4,261.13) -- (264.51,256.33) ;
\draw [shift={(269.49,255.89)}, rotate = 174.84] [fill={rgb, 255:red, 144; green, 19; blue, 254 }  ,fill opacity=1 ][line width=0.08]  [draw opacity=0] (8.57,-4.12) -- (0,0) -- (8.57,4.12) -- cycle    ;
\draw [color={rgb, 255:red, 0; green, 0; blue, 255 }  ,draw opacity=1 ][line width=2.25]    (120.99,324.43) -- (174.31,325.32) ;
\draw [shift={(179.31,325.4)}, rotate = 180.95] [fill={rgb, 255:red, 0; green, 0; blue, 255 }  ,fill opacity=1 ][line width=0.08]  [draw opacity=0] (8.57,-4.12) -- (0,0) -- (8.57,4.12) -- cycle    ;
\draw [color={rgb, 255:red, 0; green, 255; blue, 0 }  ,draw opacity=1 ][line width=2.25]    (120.99,324.43) -- (157.5,362.18) ;
\draw [shift={(160.97,365.77)}, rotate = 225.95] [fill={rgb, 255:red, 0; green, 255; blue, 0 }  ,fill opacity=1 ][line width=0.08]  [draw opacity=0] (8.57,-4.12) -- (0,0) -- (8.57,4.12) -- cycle    ;
\draw [color={rgb, 255:red, 255; green, 0; blue, 0 }  ,draw opacity=1 ][line width=2.25]    (120.99,324.43) -- (121.88,270.77) ;
\draw [shift={(121.96,265.78)}, rotate = 90.95] [fill={rgb, 255:red, 255; green, 0; blue, 0 }  ,fill opacity=1 ][line width=0.08]  [draw opacity=0] (8.57,-4.12) -- (0,0) -- (8.57,4.12) -- cycle    ;

\draw [color={rgb, 255:red, 0; green, 0; blue, 255 }  ,draw opacity=1 ][line width=2.25]    (395.49,243.91) -- (448.29,236.4) ;
\draw [shift={(453.24,235.69)}, rotate = 171.9] [fill={rgb, 255:red, 0; green, 0; blue, 255 }  ,fill opacity=1 ][line width=0.08]  [draw opacity=0] (8.57,-4.12) -- (0,0) -- (8.57,4.12) -- cycle    ;
\draw [color={rgb, 255:red, 0; green, 255; blue, 0 }  ,draw opacity=1 ][line width=2.25]    (395.49,243.91) -- (437.48,275.44) ;
\draw [shift={(441.48,278.44)}, rotate = 216.9] [fill={rgb, 255:red, 0; green, 255; blue, 0 }  ,fill opacity=1 ][line width=0.08]  [draw opacity=0] (8.57,-4.12) -- (0,0) -- (8.57,4.12) -- cycle    ;
\draw [color={rgb, 255:red, 255; green, 0; blue, 0 }  ,draw opacity=1 ][line width=2.25]    (395.49,243.91) -- (387.92,190.78) ;
\draw [shift={(387.22,185.83)}, rotate = 81.9] [fill={rgb, 255:red, 255; green, 0; blue, 0 }  ,fill opacity=1 ][line width=0.08]  [draw opacity=0] (8.57,-4.12) -- (0,0) -- (8.57,4.12) -- cycle    ;

\draw    (184.5,248) -- (206.15,258.15) ;
\draw [shift={(207.96,259)}, rotate = 205.12] [color={rgb, 255:red, 0; green, 0; blue, 0 }  ][line width=0.75]    (10.93,-3.29) .. controls (6.95,-1.4) and (3.31,-0.3) .. (0,0) .. controls (3.31,0.3) and (6.95,1.4) .. (10.93,3.29)   ;
\draw    (390.5,275.5) -- (392.75,257.98) ;
\draw [shift={(393,256)}, rotate = 97.31] [color={rgb, 255:red, 0; green, 0; blue, 0 }  ][line width=0.75]    (10.93,-3.29) .. controls (6.95,-1.4) and (3.31,-0.3) .. (0,0) .. controls (3.31,0.3) and (6.95,1.4) .. (10.93,3.29)   ;

\draw (201.93,228.27) node [anchor=north west][inner sep=0.75pt]  [rotate=-17.21]  {$\mathsf{b_{1}}$};
\draw (372.32,219.23) node [anchor=north west][inner sep=0.75pt]  [rotate=-351.62]  {$\mathsf{b_{2}}$};
\draw (100.96,306.9) node [anchor=north west][inner sep=0.75pt]    {$\mathsf{w}$};
\draw (170.12,239.58) node  [rotate=-18.58] [align=left] {\begin{minipage}[lt]{25.24pt}\setlength\topsep{0pt}
base
\end{minipage}};
\draw (344,277.5) node [anchor=north west][inner sep=0.75pt]   [align=left] {end-effector};
\draw (267,257.9) node [anchor=north west][inner sep=0.75pt]    {$\mathbf{v}$};

\end{tikzpicture}
}
  \caption{For any two reference frames $\mathsf{b_1}$ and $\mathsf{b_2}$, the rigid translation ${}^{w}{\vt}_{b_2}$ can be represented using the forward kinematics of an SP robot.}
  \label{fig:sprobot}
\end{figure}
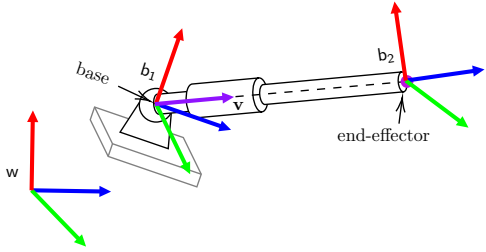

\Cref{fig:sprobot} visualizes an SP robot. The translation of the end-effector relative to the base is a function $p:\real{3}\times\real{}\times\real{3}\rightarrow \real{3}$ defined as
    \begin{equation}
        p(\vt,\tau,\vv) = \vt+\tau_u \tau \vv
    \end{equation}
    where $\vv$ is a unit vector aligned with the axis of the prismatic joint, and $\tau\in[0,1]$ is the normalized prismatic joint position. We refer to  $(\tau_i,\vv_i)$ as the \emph{pose} of the SP robot.

In the context of this paper, SP robots are a useful modeling tool because of the following property.
\begin{lemma}\label{lem:sprobot}
    Any rigid translation ${}^{w}\vt_{b_2}$ can be represented using the forward kinematics of an SP robot based at $b_1$, i.e.,
    \begin{equation}\label{eq:rigid translation and sp robot}
        {}^{w}\vt_{b_2} = p({}^{w}{\vt}_{b_1},\tau,\vv),
    \end{equation}
    where $\vv$ is the unit vector aligned with the line connecting the origins of $b_1$ and $b_2$, assuming the robot's extension cap $\tau_u\geq\norm{{}^{w}\vt_{b_2}-{}^{w}\vt_{b_1}}$.
\end{lemma}
\begin{pf}
    The unit vector is given by $\vv = \frac{{}^{w}\vt_{b_2}-{}^{w}\vt_{b_1}}{\norm{{}^{w}\vt_{b_2}-{}^{w}\vt_{b_1}}}$. Choose $\tau=\norm{{}^{w}\vt_{b_2}-{}^{w}\vt_{b_1}}/\tau_u$ and \eqref{eq:rigid translation and sp robot} holds as long as $\tau_u\geq\norm{{}^{w}\vt_{b_2}-{}^{w}\vt_{b_1}}$ (the prismatic joint is long enough).
\end{pf}

As shown in Section \ref{sec:translation relax}, the variables $(\tau,\vv)\in\mathbb{R}\times\mathcal{S}^2$ can be expressed as linear functions of $\mY_\tau$, so any linear kinematic constraints on $(\tau,\vv)$ remain linear in $\mY_\tau$, enabling convex formulations of lifted variables from convex kinematics of a virtual robot.

\subsection{Perspective-n-Point problem}
We review the perspective-n-point (PnP) problem defined below.
\begin{definition}\label{def:pnp}
    Given the global coordinates of a set of points $\{\vq_1,\dots,\vq_n\}$ and their projections $\{\vp_1,\dots,\vp_n\}$ (in pixels) onto the image plane of a camera\footnote{Throughout this work, we assume the camera has no distortion and the principal points are located at $(0,0)$ of the image plane \cite{Hartley:book04,Ma:book04}.} with the intrinsic matrix $\mK$, the PnP problem is to find the camera pose $(\mR,\vt)$ such that
    \begin{equation}\label{eq:pnp camera proj}
        \Tilde{\vp}_i = \mK(\mR\transpose\vq_i-\mR\transpose\vt),\quad \forall i=1,\dots,n,
    \end{equation}
    where $\Tilde{\vp}_i\in\real{3}$ projects to the image coordinates $\vp_i$, i.e., $\vp_i = \bmat{\Tilde{\vp}_i^{(1)}/\Tilde{\vp}_i^{(3)} &\Tilde{\vp}_i^{(2)}/\Tilde{\vp}_i^{(3)}}\transpose$. We define $\hat{\vp}_i$ as a unit vector from the camera origin passing $\vp_i$.
\end{definition}

\ifshowfullcontents
We define the following associated optimization problem.
\begin{problem}[PnP Problem]\label{prob:pnp}
    \begin{equation}\label{eq:pnp}
    \begin{aligned}
        &\min_{\mR\in\mathrm{SO}(3),\vt\in\real{3},\{\tau_i\}_n\in\real{n}} && \sum_i^n\norm{\frac{\vq_i-\vt}{\tau_i}-\mR\hat{\vp}_i}_2^2
    \end{aligned}
    \end{equation}
\end{problem}
The factor $\tau_i$ is a positive scalar. The vector $\hat{\vp}_i$ is obtained by normalizing $\bmat{\vp_i & f_{cam}}\transpose$, where $f_{cam}$ is the camera focal length. 
\else
\fi

\ifshowfullcontents
We introduce the following proposition to show that solving Problem \ref{prob:pnp} is equivalent to solving Definition \ref{def:pnp}.
\begin{proposition}\label{prop:pnp problems relation}
    A pose $(\mR^{\star},\vt^{\star})$ is a solution to the PnP problem in Definition \ref{def:pnp} if and only if there exists a $\{\tau_i^{\star}\}_n$ such that $(\mR^{\star},\vt^{\star},\{\tau_i^{\star}\}_n)$ is a minimizer of Problem \ref{prob:pnp}, and the corresponding minimal cost is zero.
\end{proposition}

\begin{pf}
    On the one hand, a minimizer $(\mR^{\star},\vt^{\star},\{\tau_i^{\star}\}_n)$ of Problem \ref{prob:pnp} yields the relation
    \begin{equation}\label{eq:pnp proof1}
       \vq_i-\vt^{\star}=\tau_i^{\star}\mR^{\star}\hat{\vp}_i.
    \end{equation}
    We substitute \eqref{eq:pnp proof1} along with $\mK=\mathrm{diag}(f_{cam}, f_{cam}, 1)$ and $\hat{\vp}_i=\bmat{\vp_i & f_{cam}}\transpose/(\norm{\bmat{\vp_i & f_{cam}}})$ to the r.h.s. of \eqref{eq:pnp camera proj}:
    \begin{equation}\label{eq:pnp proof2}
        r.h.s. = \tau_i^{\star}f_{cam}/(\norm{\bmat{\vp_i & f_{cam}}}) \bmat{\vp_i & 1}\transpose.
    \end{equation}
    It can be seen that $\vp_i$ can be recovered by dividing \eqref{eq:pnp proof2} by the third entry of itself, thus matching Definition \ref{def:pnp}.

    On the other hand, because $\vp_i$ is the projection of $\vq_i$ onto the image plane, $\vq_i-\vt$ and $\mR\vp_i$ are collinear. Moreover, because $\hat{\vp}_i$ is a normalization of $\vp_i$, $\vq_i-\vt$ and $\mR\hat{\vp}_i$ are collinear. Therefore there exists factors $\{\tau_i\}_n$ such that $\frac{\vq_i-\vt}{\tau_i}=\mR\hat{\vp}_i,\forall i=1,\dots,n$ and $(\mR,\vt,\{\tau_i\}_n)$ is a minimizer of Problem \ref{prob:pnp}.
\end{pf}

\begin{assumption}\label{as:pnp}
The following conditions hold true:
    \begin{enumerate}
        \item $n\geq6$;
        \item The points $\vq_i$ are in general position, meaning four or more of them do not lie on a single plane in $\real{3}$;
    \end{enumerate}
\end{assumption}

We then have the following regarding the uniqueness of the solution of the PnP problem.
\begin{lemma}[\cite{fischler1981random}]
    When Assumption \ref{as:pnp} holds, the PnP problem in Definition \ref{def:pnp} has a unique solution.
\end{lemma}
\fi

We define the following assembly of SP robots.
\begin{definition}
  A \emph{PnP robot} is comprised of $n$ SP robots with poses $(\tau_i,\vv_i)$, where all of their spherical joints share the same parent, which we call the base. We denote the pose of the base $({}^{w}{\tilde{\mR}}_c,{}^{w}{\tilde{\vt}}_c)$. 
\end{definition}

\subsection{Hand-eye calibration}
The robot hand-eye calibration problem is the problem of finding the transformation between the robot end effector and a rigidly mounted camera\footnote{In this paper, we consider the eye-in-hand calibration, meaning that the camera is fixed on the end-effector of the robot.}. Suppose an object with an unknown pose $(\mR_f,\vt_f)$ is represented by $n$ features and measured by the camera from multiple robot configurations indexed $i\in\{1,\dots,m\}$. We denote $\mathsf{e_i}$ and $\mathsf{c_i}$ as the frames of the end-effector and the camera under robot configuration $i$, respectively. Consider any two configurations $i$ and $i'$ as shown in Fig. \ref{fig:handeye}, the hand-eye calibration problem can be formulated using the following relation
\begin{equation}\label{eq:AXXB}
    \begin{aligned}
    \mA\mX=\mX\mB
    \end{aligned}
\end{equation}
where
\ifshowfullcontents
\begin{equation}\label{eq:A B definition}
    \begin{aligned}
        \mA &= ({}^{w}{\mT}_{e_{i'}})^{-1}{}^{w}{\mT}_{e_i}, \text{and }\\
        \mB &= {}^{c_{i'}}\mT_{f}({}^{c_i}\mT_{f})^{-1}.
    \end{aligned}
\end{equation}
Transformations ${}^{w}{\mT}_{e_2},{}^{w}{\mT}_{e_1}$ are computed by performing robot calibration and ${}^{c_2}\mT_{f},{}^{c_1}\mT_{f}$ are found from camera calibration. The goal of hand-eye calibration is to find the unknown transformation matrix $\mX$ as a solution to \eqref{eq:AXXB}.
\fi
The transformation $\mX$ can be decomposed into $\mR_x\in\mathrm{SO}(3)$ and $\vt_x\in\real{3}$.

We define a virtual robot for the hand-eye calibration task by combining SP and PnP robots.
\begin{definition}
    A hand-eye robot has $m$ bases, where the transformation from each base $i\in\{1,\dots,m\}$ to the world is ${}^{w}{\mT}_{e_i}$. From each base, an SP robot connects to the camera origin of $\mathsf{c_i}$. A PnP robot connects to all features $\{f_j\}_n$ and the base of the PnP robot is attached to $\mathsf{c_i}$. We denote the sets of all poses of the hand-eye robot as $(\tauset,\cV)$.
\end{definition}
A visualization of the hand-eye robot is in Fig. \ref{fig:hand-eye-kinematic-chain}.


\subsection{Simultaneous calibration of dual-robot systems}
We consider another calibration scenario shown in Fig. \ref{fig:axbycz}, where the hand-eye, hand-target, and robot-robot transformations are calibrated simultaneously. The coordinate systems $\mathsf{ea}$, $\mathsf{eb}$, $\mathsf{c}$, $\mathsf{t}$, and $\mathsf{b}$ are rigidly attached to the end-effector of robot 1, the end-effector of robot 2, the camera attached to robot 1, the calibration target attached to robot 2, and the base of robot 2, respectively. The relations among these coordinate systems satisfy the equation
\begin{equation}
    \mA\mX\mB=\mY\mC\mZ, \label{eq:axb_eq_ycz}
\end{equation}
where $\mX$, $\mY$, and $\mZ$ are the unknown transformation matrices 
\ifshowfullcontents
\begin{equation}\label{eq:axycb-xyz_def}
    \mX = \mT_{ea}^{-1}\mT_{c}, \quad \mY = \mT_{b}, \quad \mZ = \mT_{eb}^{-1}\mT_{t}.
\end{equation}
The matrices $\mA$, $\mB$, and $\mC$ are the measured transformations
\begin{equation}
    \mA = {}^w\mT_{ea}, \quad \mB = {}^c\mT_{t}, \quad \mC = {}^{b}\mT_{eb}.
\end{equation}
\fi

We propose the following virtual robot for the simultaneous calibration of dual-robot systems.

\begin{definition}
    A dual calibration robot has three fixed links, parts \robotpart{A}, \robotpart{B}, and \robotpart{C}. Reference frames are attached to the ends of the three parts. The transformation between the two attached reference frames are fixed to the values $\mA$, $\mB$, and $\mC$ for \robotpart{A}, \robotpart{B}, and \robotpart{C}, respectively. Each pair of the parts is connected via a spherical joint followed by a prismatic joint, then another spherical joint, as shown in \Cref{fig:dual-hand-eye-robot}. The lower end of \robotpart{A} is rigidly attached to the world reference frame. 
    The dual-robot system may have different configurations, where $m$ measurements $\{\mA_i,\mB_i,\mC_i\}_m$ are taken.
\end{definition}

\ifshowfullcontents
The coordinate systems in the dual-robot system can be represented by the reference frames of a dual calibration robot. The parameters $\mA$, $\mB$, and $\mC$ are modeled using rigid bodies, while the unknown transformations $\mX$, $\mY$, and $\mZ$ are represented using SP robots. The unknown transformations $\mX,\mY,\mZ$ are assumed to remain the same among different configurations, i.e.,
\begin{equation}\label{eq:xyz consistency}
    \mX_i = \mX_{i'},\mZ_i = \mZ_{i'},\text{ and } \mY_i = \mY_{i'},\quad \forall i,i'=1,\dots,m.
\end{equation}
Such equality of transformations will be discussed in detail in Section \ref{sec:kine probs}.
\fi

\subsection{Kinematics problems}\label{sec:kine probs}
\begin{table}[ht]
    \centering
    \vspace{-5pt}
    \begin{tabular}{|c||c|c|c|}
    \hline
       Problem & PnP & Hand-Eye & Dual Calibration\\
       \hline
       \hline
       \makecell{\scriptsize Number of \\SP robots} & $n$ & $m(n+1)$ & $3m$ \\
       \hline
       \makecell{\scriptsize Number of \\unknown \\Rotations} & $1$ & $m+1$ & $3m$ \\
       \hline
       \makecell{\scriptsize Objective \\Function $f_k$} & Reproj. Error & Reproj. Error & {\tiny$\norm{\mA\mX\mB-\mY\mC\mZ}_{F}$}\\
       \hline
       Constraints & Kine. Closure & \makecell{Kine. Closure, \\$\mX$ Equality} &$\mX,\mY,\mZ$ Equality\\
       \hline
    \end{tabular}
    \caption{Kinematics problems of the virtual robots}
    \label{tab:kine-probs}
\end{table}

We define the following kinematics problem for the PnP, hand-eye, and dual calibration robots. See Table \ref{tab:kine-probs} for a summary of the problem settings for specific problems.
\begin{problem}[Kinematics Problem of Virtual Robots]\label{prob:general kine prob}
    \begin{subequations}\label{eq:general kine prob}
        \begin{align}
            &\min_{\tauset,\cV,\cR} && f_k(\tauset,\cV,\cR)\\
        &\text{s.t.} && e_{kc}(\tauset,\cV,\cR)=0\label{eq:gen-kine-close}\\
        & && e_{te}(\tauset,\cV,\cR)=0\label{eq:gen-fixed-transform}
        \end{align}
    \end{subequations}
\end{problem}
The variable $\{\tauset,\cV\}$ is the set of poses of SP robots. The variable $\cR$ is the set of unknown rotation matrices. The functions $f_k$, $e_{kc}$ and $e_{te}$ stand for the \emph{kinematics objective}, \emph{kinematic closure} and \emph{transformation equality} functions, respectively, which we discuss in detail below with example.

\textbf{\textit{Objective function.}} The objective functions encode the estimation and calibration metrics. For the PnP and hand–eye robots, the objective is the reprojection error, defined as the norm of the difference between the prismatic joint unit vectors $\vv_i$ and the oriented measurements $\mR_c\hat{\vp}_i$. For the dual calibration robot, the objective is given by $\norm{\mA\mX\mB-\mY\mC\mZ}_F$.

\textit{\textbf{Constraints.}} The kinematic constraints of the following types are defined to restrict the feasible set so that any solution to the kinematics problem is also a feasible solution to the original estimation or calibration problem.
\begin{itemize}
    \item \textbf{Kinematic closure.} This constraint restricts the kinematic closure of the kinematic chains.
    \item \textbf{Transformation Equality.} This constraint restricts the consistency of some transformations in different configurations. For example, $\mX_i=\mX_{i'}$ for any two measurements $i\ne i'$.
\end{itemize}

\begin{example}[PnP robot kinematics]
The following proposition relates the kinematics problems of the PnP robot and the PnP problem.
\begin{proposition}\label{prop:pnprobot}
    Let $\{\vq_1,\dots,\vq_n\}$ be a set of points in general positions and $n\geq6$, and let $\{\vp_1,\dots,\vp_n\}$ be their projections onto the image plane of a camera with intrinsics $\mK$ and pose $({}^{w}{\mR}_c,{}^{w}{\vt}_c)$. We attach every end-effector of a PnP robot to $\vq_i$. Then $({}^{w}{\tilde{\mR}}_c,{}^{w}{\tilde{\vt}}_c) = ({}^{w}{\mR}_c,{}^{w}{\vt}_c)$ if and only if $(\{\vv_i,\tau_i\}_n,{}^{w}{\tilde{\mR}}_c,{}^{w}{\tilde{\vt}}_c)$ is a minimizer of
    \begin{problem}[PnP Robot Kinematics]\label{prob:pnprobot}
    \begin{subequations}\label{eq:pnprobot}
        \begin{align}
            &\min_{\{\vv_i,\tau_i\}_n,{}^{w}{\tilde{\mR}}_c,{}^{w}{\tilde{\vt}}_c} && f_k:=\sum_i^n\norm{\vv_i-{}^{w}{\tilde{\mR}}_c\hat{\vp}_i}_2^2\label{eq:pnp robot obj}\\
        &\text{s.t.} && p({}^{w}{\tilde{\vt}}_{c},\tau_i,\vv_i) = \vq_i, \forall i=1,\dots,n\label{eq:pnp robot constraint}
        \end{align}
    \end{subequations}
\end{problem}
and the corresponding minimal cost is zero.
\end{proposition}
\begin{pf}
    On the one hand, when $(\{\vv_i,\tau_i\}_n,{}^{w}{\tilde{\mR}_c},{}^{w}{\tilde{\vt}}_c)$ minimizes the objective function and the corresponding cost is zero, we have $\vv_i={}^{w}{\tilde{\mR}}_c\hat{\vp}_i$ and ${}^{w}{\tilde{\vt}}_c+\tau_u\tau_i\vv_i=\vq_i$, $\forall i=1,\dots,n$. This means that $\frac{\vq_i-{}^{w}{\tilde{\vt}}_c}{\tau_u\tau_i}={}^{w}{\tilde{\mR}}_c\hat{\vp}_i$. By Proposition \ref{prop:pnp problems relation}, we know that $({}^{w}{\tilde{\mR}_c},{}^{w}{\tilde{\vt}}_c)$ is a solution to the PnP problem and therefore $({}^{w}{\tilde{\mR}}_c,{}^{w}{\tilde{\vt}}_c) = ({}^{w}{\mR}_c,{}^{w}{\vt}_c)$.

    On the other hand, when $({}^{w}{\tilde{\mR}}_c,{}^{w}{\tilde{\vt}}_c) = ({}^{w}{\mR}_c,{}^{w}{\vt}_c)$, by Proposition \ref{prop:pnp problems relation}, there exists a $\{\tilde{\tau}_i\}_n$ such that $\frac{\vq_i-{}^{w}{\tilde{\vt}}_c}{\tilde{\tau}_i}={}^{w}{\tilde{\mR}}_c\hat{\vp}_i, \forall i=1,\dots,n$. Let $\tilde{\tau}_i = \tau_u\tau_i$ and $\vv_i=\frac{\vq_i-{}^{w}{\tilde{\vt}}_c}{\tilde{\tau}_i}$ and $(\{\vv_i,\tau_i\}_n,{}^{w}{\tilde{\mR}}_c,{}^{w}{\tilde{\vt}}_c)$ is a minimizer of Problem \ref{prob:pnprobot} with minimal cost equals to zero.
\end{pf}

Problem $\ref{prob:pnprobot}$ minimizes the reprojection error \eqref{eq:pnp robot obj} while constraining the kinematic closure through \eqref{eq:pnp robot constraint}. Proposition \ref{prop:pnprobot} enables us to convert a PnP problem to a kinematics problem of a virtual PnP robot.
\end{example}

\ifshowfullcontents
\begin{example}[Hand-eye robot Kinematic constraints]
    We investigate the hand-eye robot to demonstrate the kinematics constraints. Observe that \eqref{eq:AXXB} is equivalent to the following equation by substituting \eqref{eq:A B definition}
\begin{equation}\label{eq:AXXB_fk}
    \begin{aligned}
    {}^w{\mT}_{e_i}\mX{}^{c_i}\mT_{f}={}^w{\mT}_{e_{i'}}\mX{}^{c_{i'}}\mT_{f}
    \end{aligned}
\end{equation}
It can be seen that \eqref{eq:AXXB_fk} encodes two constraints $\forall i,i'\in\{1,\dots,m\}$:
    \begin{subequations}\label{eq:AXXB_break_down}
        {\footnotesize
        \begin{align}
            &\text{Kinematics closure: }{}^{w}{\mT}_{e_i}{}^{e_i}{\mT}_{c_i}{}^{c_{i}}\mT_f={}^{w}{\mT}_{e_{i'}}{}^{e_{i'}}{\mT}_{c_{i'}}{}^{c_{i'}}\mT_f\label{eq:AXXB_break_down_1}\\
            &\text{Transformation equality: }\mX = {}^{e_i}{\mT}_{c_i}={}^{e_{i'}}{\mT}_{c_{i'}}.\label{eq:AXXB_break_down_2}
        \end{align}}
    \end{subequations}
The \eqref{eq:AXXB_break_down_1} is a kinematic closure constraint, meaning that the transformations from the target to the world frame are the same following the kinematic chain of every measurement: $\mathsf{f}\rightarrow \mathsf{c_i}\rightarrow\mathsf{e_i}$. Constraint \eqref{eq:AXXB_break_down_2} enforces that the transformations $\{{}^{e_i}{\mT}_{c_i}\}_m$ equals $\mX$ for every measurement.
\end{example}
\fi

\subsection{Semi-Definite Relaxation}
Consider the variable set
\begin{equation}
    \cX:=\{\tauset,\cV,\cR,\tauset\cV\}\label{eq:vxr}
\end{equation}
The set $\tauset\cV$ is the product set $\tauset\cV:=\{\tau\vv|\tau\in\tauset,\vv\in\cV\}$.
In general, for the defined virtual robots, the kinematics Problem \ref{prob:general kine prob} consists of \begin{enumerate*}[label=(\roman*)]
    \item an objective function $f_k$ that is quadratic to $\cX$, and \item constraints linear to $\cX$. This means Problem \ref{prob:general kine prob} is convex in $\cX$ (see Problem \ref{prob:pnprobot} for an example).
\end{enumerate*}

As we have shown in Section \ref{sec:so3 relax} and \ref{sec:translation relax}, $\cR$, and $\tauset,\cV,\tauset\cV$ can be represented exactly using rank-1 matrices $\mY=\{\mY_i|\mY_i\in\cY\}$, $\mY_\tau=\{\{\mY_\tau\}_3|\{\mY_\tau\}_3\in\cY_\tau\}$ through linear functions $g(\mY)$, $g_\tau(\mY_\tau)$, and $g_{\tau v}(\mY_\tau)$. Using this, we propose Algorithm \ref{alg:solve_sdp} for solving $\cX$.
\begin{algorithm}[h]
\caption{Solution procedure for Problem~\ref{prob:general kine prob}}
\label{alg:solve_sdp}
\begin{algorithmic}[1]
    \State Develop Problem~\ref{prob:sdp-relax} from Problem~\ref{prob:general kine prob} by representing $\cX$ using linear functions of $\mY,\mY_\tau$.
    \State Solve Problem~\ref{prob:sdp-relax} to obtain an initial feasible solution $(\mY^0,\mY_\tau^0)$ not necessarily rank-1.
    \State Move $(\mY^0,\mY_\tau^0)$ to a rank-1 solution $(\mY^r,\mY_\tau^r)$ by iteratively solving Problem~\ref{prob:simple rank min update}.
    \State Move $(\mY^r,\mY_\tau^r)$ by iteratively solving sequences of Problem~\ref{prob:low_rank_channel} to a low-cost solution $(\hat{\mY},\hat{\mY}_\tau)$ up to a tolerance $\epsilon$, i.e., $f(\hat{\mY},\hat{\mY}_\tau)<\epsilon$. \label{step:cost_improve}
    \State Certify the optimality by solving the dual Problem~\ref{prob:dual} and checking the duality gap.
\end{algorithmic}
\end{algorithm}

\begin{remark}
    We remark that, for the dual hand-eye robot, $\cX$ needs to include $\cV\cR,\cR\cR$ for Problem \ref{prob:general kine prob} to be convex.
    \ifshowfullcontents
    We will discuss how to account for these additional variables in Section \ref{sec:addtnl variables}.
    \fi
\end{remark}

\ifshowfullcontents
\section{Kinematics Problems of Virtual Robots}
In this section, we first introduce a general formulation of the transformation equality constraint. Then, we investigate the kinematics problems of the hand-eye and dual calibration robots. 
\subsection{Constraint on constant transformation between two reference frames}
\begin{figure}[H]
  \centering
  \resizebox{0.38\textwidth}{!}{\input{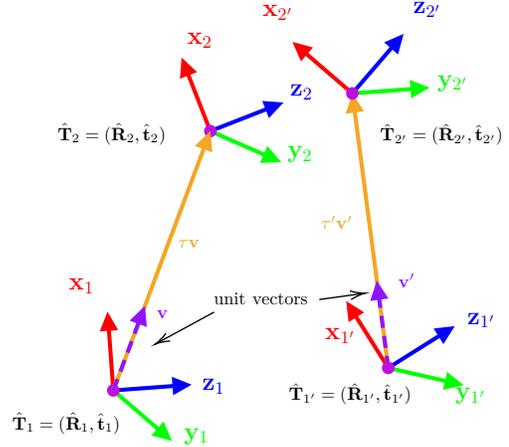}}
  \caption{The consistency of transformations between reference frames with unknown absolute poses, $\mT_{1}^{-1}\mT_{2} = \mT_{1'}^{-1}\mT_{2'}$, can be enforced through constraints on $\hat{\mR}_1$, $\hat{\mR}_2$, $\hat{\mR}_{1'}$, $\hat{\mR}_{2'}$, $\tau$, $\tau'$, $\vv$, and $\vv'$.}
  \label{fig:hand-eye-transform}
\end{figure}


Consider two reference frames $\mathsf{f_1},\mathsf{f_2}$ and their corresponding frames after some rigid-body transformations, $\mathsf{f'_1},\mathsf{f'_2}$. We denote the global poses $\hat{\mT}_1 := {}^{w}\mT_{f_1} = (\hat{\mR}_1,\hat{\vt}_1)$ and $\hat{\mT}_2 := {}^{w}\mT_{f_2} = (\hat{\mR}_2,\hat{\vt}_2)$ (and similarly, $\hat{\mT}_{1'}$ and $\hat{\mT}_{2'}$). Suppose the two frames are attached rigidly and the relative transformation remains the same, i.e., ${}^{f_1}\mT_{f_2}={}^{f_{1'}}\mT_{f_{2'}}$, which is equivalent to

\begin{equation}\label{eq:constant_rel_transform}
    \hat{\mT}_{1}^{-1}\hat{\mT}_{2} {=} \hat{\mT}_{1'}^{-1}\hat{\mT}_{2'}.
\end{equation}
The decoupled translations and rotations of \eqref{eq:constant_rel_transform} satisfy
\begin{subequations}\label{eq:constant transform}
\begin{align}
    \hat{\mR}_{1}\transpose(\hat{\vt}_{2}{-}\hat{\vt}_{1})&{=}\hat{\mR}_{1'}\transpose(\hat{\vt}_{2'}{-}\hat{\vt}_{1'})\label{eq:constant rel translation}\\
    \hat{\mR}_1\transpose\hat{\mR}_2&{=}\hat{\mR}_{1'}\transpose\hat{\mR}_{2'}.\label{eq:constant rel rotation}
\end{align}
\end{subequations}
Recall that, by Lemma \ref{lem:sprobot}, $\hat{\vt}_2$ can be represented using the forward kinematics of an SP robot connecting the frames $1$ and $2$. We denote the prismatic variables of this SP robot as $\tau,\vv$ and $\tau',\vv'$ between the motion, then substitute $\hat{\vt}_2 = p(\hat{\vt}_1,\tau,\vv)$, $\hat{\vt}_{2'} = p(\hat{\vt}_1,\tau',\vv')$ into the \eqref{eq:constant rel translation} and get
\begin{equation}\label{eq:transform equality r tau v}
    \hat{\mR}_1\transpose\tau\vv = \hat{\mR}_{1'}\transpose\tau'\vv'.
\end{equation}

\begin{definition}
    For two reference frame $1$ and $2$, the transformation equality constraint is defined as $\{\tau,\vv,\hat{\mR}_1,\hat{\mR}_2,\tau',\vv',\hat{\mR}_{1'},\hat{\mR}_{2'}\}\in\cC_{ft}$ such that
    \begin{subequations}\label{eq:transform fix}
    \begin{align}
        \tau &= \tau'.\label{eq:extension fix}\\
       \hat{\mR}^{(l_1)}_1{}\transpose\vv &=\hat{\mR}^{(l_1)}_{1'}{}\transpose\vv',\forall l_1 = 1,2,3,\text{ and,}\label{eq:angle fix}\\
        \hat{\mR}^{(l_1)}_1{}\transpose\hat{\mR}_2^{(l_2)}&=\hat{\mR}^{(l_1)}_{1'}{}\transpose\hat{\mR}_{2'}^{(l_2)}, \forall l_1\in\{1,2,3\}, l_2\in\{1,2\}.\label{eq:rotation fix}
    \end{align}
\end{subequations}
The indices $\{l_1,l_2\}$ are for the columns of the rotation matrices $\mR_1$ and $\mR_2$.
\end{definition}

\begin{proposition}\label{prop:constant transformation hand-eye}
    The relative transformation between two reference frames $1$ and $2$ remains the same, i.e., \eqref{eq:constant transform} holds, if and only if $\{\tau,\vv,\hat{\mR}_1,\hat{\mR}_2,\tau',\vv',\hat{\mR}_{1'},\hat{\mR}_{2'}\}\in\cC_{ft}$.
    
\end{proposition}
\begin{pf}
    We start with the relative translation \eqref{eq:constant rel translation} by multiplying $\tau$ and $\tau'$ on left and right of \eqref{eq:angle fix} and concatenate for all of the columns $i$ we obtain $\hat{\mR}_{1}\transpose(\hat{\vt}_{2}{-}\hat{\vt}_{1})=\hat{\mR}_{1'}\transpose(\hat{\vt}_{2'}{-}\hat{\vt}_{1'})$.
    
    We now prove for the relative rotation \eqref{eq:constant rel rotation}. Rotation preserves cross product, meaning that 
    \begin{equation}
        \hat{\mR}^{(i)}_1{}\transpose(\hat{\mR}_2^{(1)}\times\hat{\mR}_2^{(2)}) = (\hat{\mR}^{(i)}_1{}\transpose\hat{\mR}_2^{(1)})\times(\hat{\mR}^{(i)}_1{}\transpose\hat{\mR}_2^{(2)}).
    \end{equation}
    Using this result and \eqref{eq:rotation fix}, we can obtain
   \begin{equation}
       \hat{\mR}^{(i)}_1{}\transpose(\hat{\mR}_2^{(1)}\times\hat{\mR}_2^{(2)})=\hat{\mR}^{(i)}_{1'}{}\transpose(\hat{\mR}_{2'}^{(1)}\times\hat{\mR}_{2'}^{(2)}), \forall i\in\{1,2,3\}.\label{eq:rotation fix proof1}
   \end{equation}
   Because $\hat{\mR}_2^{(3)}=\hat{\mR}_2^{(1)}\times\hat{\mR}_2^{(2)}$, we have
   \begin{equation}
       \hat{\mR}^{(i)}_1{}\transpose\hat{\mR}_2^{(3)}=\hat{\mR}^{(i)}_{1'}{}\transpose\hat{\mR}_{2'}^{(3)}, \forall i\in\{1,2,3\}.\label{eq:rotation fix proof2}
   \end{equation}
   Then \eqref{eq:constant rel rotation} is a concatenation of \eqref{eq:rotation fix} and \eqref{eq:rotation fix proof2} for $(i,j)\in\{1,2,3\}$.
\end{pf}

Intuitively, the reference frames are connected using SP robots, and the equality of the relative transformation can be bounded by \eqref{eq:transform fix} acting on the variables related to the SP robots.

\begin{remark}
    For the dual calibration robot, the constraints \eqref{eq:angle fix} and \eqref{eq:rotation fix} can be
    \begin{itemize}
        \item \emph{linear} when $\mR_1$ is known and $\mR_2$ is unknown (e.g., $\mX$ and $\mY$);
        \item \emph{quadratic} when $\mR_1$ $\mR_2$ are both unknown (e.g., $\mZ$).
    \end{itemize}
     For the latter case to fit in a convex problem, we will discuss how to relax these into linear constraints in Section \ref{sec:addtnl variables}.
\end{remark}

\subsection{Hand-eye robot}

The hand-eye robot is designed to model the transformations within the hand-eye calibration problem using the robot's poses. Specifically, we have fixed the poses between the hand-eye robot's bases to be $\{{}^{w}{\mT}_{e_i}\}_m$. Moreover, we want to represent the camera poses $\{{}^{w}{\mT}_{c_i}\}_m$ using the pose of the PnP robots' bases $\{{}^{w}{\tilde{\mR}}_{c_i},{}^{w}{\tilde{\vt}}_{c_i}\}_n$. Once these equivalences are established, we can find the solution to the hand-eye calibration problem $\mX$ by computing the relative transformation of the poses of $\mathsf{e_i}$ and $\mathsf{c_i}$.
We will show that this can be done using proper constraints.

To begin with, we define the unknown variables that, once solved in the kinematics problem, can be used to recover the transformation $\mX$.

\begin{equation}\label{eq:hand-eye-variables}
    \begin{aligned}
        \{\mR\}&:=\{\mR_{c_i}\}_m,\mR_{f}\\
    \tauset&:=\{\tau_{e_i,c_i}\}_m, \{\tau_{c_i,f_j}\}_{i=1,\dots,m,j=1,\dots,n},\text{ and}\\
    \cV&:=\{\vv_{e_i,c_i}\}_m,\{\vv_{c_i,f_j}\}_{i=1,\dots,m,j=1,\dots,n},
    \end{aligned}
\end{equation}
The rotations include all the camera orientations and the calibration object. The subscripts like $\{e_i,c_i\}$ mean that the prismatic variable is for the connection $e_i$ and $c_i$. For brevity, we rename $\tau_{0i}:=\tau_{e_i,c_i}$, $\vv_{0i}:=\vv_{e_i,c_i}$, $\tau_{ij}:=\tau_{c_i,f_j}$, and $\vv_{ij}:=\vv_{c_i,f_j}$.

So far, we have derived the constraints in the form of \eqref{eq:AXXB_break_down}. We first look at \eqref{eq:AXXB_break_down_1}, and define the forward kinematics of a chain: $\mathsf{e_i}\rightarrow \mathsf{c_i}\rightarrow f_j\rightarrow\mathsf{f}$. 
\begin{equation}\label{eq:handeyerobot_kinematics_close1}
    p_{fk,ij}(\tauset,\cV,{}^w\mR_{f}):=p(p({}^{w}{\vt}_{e_i},\tau_{0i},\vv_{0i}),\tau_{ij},\vv_{ij}) -{}^{w}{\mR}_f\vf_j
\end{equation}
The vector $\vf_j$ is the feature position relative to the feature frame, available before the calibration.
Next, we equate \eqref{eq:handeyerobot_kinematics_close1} for different $(i,j)$ to get
\begin{equation}\label{eq:kinematic_close_constraint}
    \begin{aligned}
        \forall i,i' = 1,\dots,m, &\quad j,j'=1,\dots,n:\\
    p_{fk,ij}(\tauset,\cV,{}^w\mR_{f})&=p_{fk,i'j'}(\tau,\cV,{}^w\mR_{f})
    \end{aligned}
\end{equation}
Observe that \eqref{eq:kinematic_close_constraint} is the translation part of \eqref{eq:AXXB_break_down_1}. We note that the rotation part for \eqref{eq:AXXB_break_down_1} is trivial as the target 

Next, we decouple \eqref{eq:AXXB_break_down_2} into rotation and translation then substitute $\vt_{c_i}=\vt_{e_i}+\mR_{e_i}{}^{e_i}\vt_{c_i}$ and get
\begin{subequations}
    \begin{align}
        \mR_{e_{i}}\transpose\mR_{c_{i}}&=\mR_{e_{i'}}\transpose\mR_{c_{i'}}\label{eq:hand-eye rotation}\\
        \mR_{e_{i}}\transpose\vt_{c_{i}}{-}\mR_{e_{i'}}\transpose\vt_{c_{i'}}&=\mR_{e_{i}}\transpose\vt_{e_{i}}{-}\mR_{e_{i'}}\transpose\vt_{e_{i'}}\label{eq:hand-eye translation0}
    \end{align}
\end{subequations}

Then we substitute $\vt_{c_{i}}=p(\vt_{e_i},\tau_{0i},\vv_{0i})$ into \eqref{eq:hand-eye translation0} and
\begin{equation}
     \mR_{e_{i}}\transpose \tau_{0i}\vv_{0i}=\mR_{e_{i'}}\transpose\tau_{0i'}\vv_{0i'}.\label{eq:hand-eye translation}
\end{equation}
Observe that \eqref{eq:hand-eye translation} is exactly \eqref{eq:transform equality r tau v}. Now, we summarize the objective function and constraints
\begin{equation}\label{eq:handeye-prob-setting}
    \begin{aligned}
        &h_k(\tau,\cV,\cR):=f_1(\cV,\mR_{c_i}) = \sum_i^m\sum_j^n\norm{\vv_{ij}-\mR_{c_i}\hat{\vp}_i}_2^2\\
        &e_{kc}(\tauset,\cV,\cR)=0: \quad \eqref{eq:kinematic_close_constraint}\\
        &e_{te}(\tauset,\cV,\cR)=0: \quad \eqref{eq:hand-eye rotation},\eqref{eq:hand-eye translation}
    \end{aligned}
\end{equation}

\begin{proposition}
    When the features satisfy Assumption \ref{as:pnp}, if there exists a solution $(\tauset^*,\cV^*,\mR^*_{c_i},\mR^*_f)$ to Problem \ref{prob:general kine prob} using the setting \eqref{eq:handeye-prob-setting} and $f_1(\cV^*,\mR^*_{c_i})=0$, then the camera poses $\{{}^{w}{}\hat{\mT}_{c_i}\}_m:=\{(\mR^*_{c_i},p(\vt^*_{e_i},\tau^*_{0i},\vv^*_{0i}))\}_m$ equals the camera poses $\{{}^{w}{}\mT_{c_i}\}_m$ in the hand-eye calibration problem and $\mX = {}^{w}{}\mT_{e_i}\transpose{}^{w}{}\hat{\mT}_{c_i}$.
\end{proposition}
\begin{pf}
    By Proposition \ref{prop:pnprobot} we know that the camera pose ${}^{w}{}\hat{\mT}_{c_i}=(\hat{\mR}_{c_i},\hat{\vt}_{c_i})$ is unique for the measurement of each camera $c_i$. Thus ${}^{w}{}\hat{\mT}_{c_i} = {}^{w}{}\mT_{c_i}$ because otherwise the uniqueness would be violated. 
    Because $\hat{\vt}_{c_i}$ is fixed, the SP robot between $e_i$ and $c_i$ is fixed, meaning we can express $\hat{\vt}_{c_i}$ using the forward kinematics of this SP robot $p(\vt^*_{e_i},\tau^*_{0i},\vv^*_{0i}))$. Finally, $\mX$ is derived as the relative transformation between ${}^{w}{}\mT_{e_i}$ and ${}^{w}{}\hat{\mT}_{c_i}$
\end{pf}

\subsection{Dual calibration robot}
To represent the configuration of the dual calibration robot, we need the following variables 
\begin{equation}\label{eq:dual-hand-eye-variables}
    \begin{aligned}
        \{\mR\}&:=\{\mR_{c_i}, \mR_{eb_i}, \mR_{t_i}\}_m,\\
    \tauset&:=\{\tau_{ea,c,i},\tau_{eb,t,i},\tau_{w,b,i}\}_m,\text{ and}\\
    \cV&:=\{\vv_{ea,c,i},\vv_{eb,t,i},\vv_{w,b,i}\}_m,
    \end{aligned}
\end{equation}
where $\mR$ are the rotation matrices corresponding to the reference frames of the camera and the end-effector of the second robot, $\tauset,\cV$ are the prismatic variables for all of the SP robots in the dual calibration robot. These variables uniquely represent the configuration of the dual calibration robot in the sense that the pose of every reference frame on the robot can be represented as a function of \eqref{eq:dual-hand-eye-variables} exclusively.

The goal of the calibration is to minimize 
\begin{equation}\label{eq:axb-ycz}
    \hat{h}_2 := \sum_{i=1}^{m}\norm{\mA_i\mX_i\mB_i-\mY_i\mC_i\mZ_i}_F^2.
\end{equation}
Observe that, for each sample $i$, \eqref{eq:axb-ycz} is the difference of the transformations obtained using the forward kinematics of two sub-trees of the kinematics chain of the dual calibration robot from the base to the target $t$. We develop the following cost function to model this difference.

\begin{equation}
    f_2:=\sum_{i=1}^m\left(\norm{\mR_{c_i}\mR_{B_i}-\mR_{t_i}}^2_F{+}\gamma\bar{p}_{fk}^2(\{\tauset,\cV,\mR\})\right)
\end{equation}

The symbol $\gamma$ is a weight to balance the terms. The function $\bar{p}_{fk}$ is the difference of the translations following the forward kinematics, defined as
\begin{equation}
\begin{aligned}
    \bar{p}_{fk}^2&:= \norm{p_{fk1} - p_{fk2}}_2^2, \text{ where}\\
    p_{fk1}&:= p(\vt_{ea_i},\tau_{ea,c,i},\vv_{ea,c,i})+\mR_{c_i}\vt_{\mB_i},\\ p_{fk2}&:=p(p(\mathbf{0},\tau_{w,b,i},\vv_{w,b,i})+\mR_{eb_i}\mR_{\mC_i}\transpose\vt_{\mC_i},\tau_{eb,t,i},\vv_{eb,t,i}).
\end{aligned}
\end{equation}

We define the following kinematic problem for the calibration task.

\begin{problem}\label{prob:dualhandeyekinematics}
{\footnotesize
    \begin{subequations}\label{eq:dualhandeyekinematics}
        \begin{align}
        &\min_{\tauset,\cV,\{\mR\}} f_2(\tauset,\cV,\{\mR\}) \\
        &\subjectto  
        \quad\forall i,i' = 1,\dots,m:\notag\\
        &\{\tau_{ea,c,i}{,}\vv_{ea,c,i},\mR_{ea_i}{,}\mR_{c_i},\tau_{ea,c,i'}{,}\vv_{ea,c,i'},\mR_{ea_{i'}},\mR_{c_{i'}}\}{\in}\cC_{ft}\label{eq:fix X}\\
        &\{\tau_{w,b,i},\vv_{w,b,i},\mR_{w_i},\mR_{b_i},\tau_{w,b,i'},\vv_{w,b,i'},\mR_{w_{i'}},\mR_{b_{i'}}\}{\in}\cC_{ft}\label{eq:fix Y}\\
        &\{\tau_{eb,t,i},\vv_{eb,t,i},\mR_{eb_i},\mR_{t_i},\tau_{eb,t,i'},\vv_{eb,t,i'},\mR_{eb_{i'}},\mR_{t_{i'}}\}{\in}\cC_{ft}\label{eq:fix Z}
        \end{align}
    \end{subequations}
    }
\end{problem}
Problem \ref{prob:dualhandeyekinematics} minimizes the calibration goal \eqref{eq:axb-ycz} while enforcing the assumption \eqref{eq:xyz consistency} through the constraints \eqref{eq:fix X} - \eqref{eq:fix Z}.

\subsection{Additional variables for constant transformations}\label{sec:addtnl variables}
To account for the quadratic constraints \eqref{eq:angle fix} and \eqref{eq:rotation fix}, we introduce the following variables.
{\scriptsize
\begin{equation}\label{eq:ya definition}
\begin{aligned}
    \mY&_a := \{\mY_{a,i}\}_m\text{, where}\\
    \mY&_{a,i} := \\
    &\{\bmat{\hat{\mR}_1^{(l_1)}\\\vv\\1}\bmat{\hat{\mR}_1^{(l_1)}\\\vv\\1}\transpose\}_{l_1=1,2,3}\cup\{\bmat{\hat{\mR}_1^{(l_1)}\\\hat{\mR}_2^{(l_2)}\\1}\bmat{\hat{\mR}_1^{(l_1)}\\\hat{\mR}_2^{(l_2)}\\1}\transpose\}_{l_1=1,2,3, l_2=1,2}
\end{aligned}
\end{equation}
}
The variable $\mY_{a,i}$ is a set of 9 7x7 matrices associated with the $i$-th configuration. Similar to $\mY_i$, each matrix in $\mY_{a,i}$ is also fixed-trace so long as $\hat{\mR}_1,\hat{\mR}_2\in \mathrm{SO}(3)$ and $\vv\in\cS^2$. For the dual calibration robot, the only unknown transformation of two reference frames with unknown poses is $\mZ$. Explicitly, $\{\tau,\vv,\hat{\mR}_1,\hat{\mR}_2,\tau',\vv',\hat{\mR}_{1'},\hat{\mR}_{2'}\}$ is linked to the robot as follows.
\begin{equation}\label{eq:ya-link-to-robot}
\begin{aligned}
    &\hat{\mR}_1=\mR_{eb_i},\hat{\mR}_2=\mR_{t_i} \text{ and }\\
    &\tau=\tau_{eb,t,i}, \vv=\vv_{eb,t,i}, \forall i=1,\dots,m.
\end{aligned}
\end{equation}
We define the following constraints to link $\mY_a$ with these variables.
\begin{definition}\label{def:c av define}
    The additional-variable constraint $\mY,\mY_{\tau},\mY_a\in\cC_{av}$ is defined such that 
    \begin{enumerate}
        \item \eqref{eq:ya-link-to-robot} holds;
        \item $\{\mM\in\cY|\forall \mM\in\mY_{a,i}\},\forall i=1,\dots,m$.
    \end{enumerate}
\end{definition}

As mentioned before, $\mR_{eb_i}, \mR_{t_i}$, and $\vv_{eb,t,i}$ can be written as linear functions $g(\mY)$ and $g_\tau(\mY_{\tau})$. Item \textit{1)} of Definition \ref{def:c av define} is enforced by having the r.h.s of \eqref{eq:ya-link-to-robot} represented through these linear functions. Item \textit{2)} is to restrict the internal structure of $\mY_{a,i}$ through the same constraint $\cY$ as $\mY_i$ for they have a similar structure of outer products of stacked unit vectors.
\fi

\vspace{-5pt}
\section{Simulation Results}\label{sec:sim}

\subsection{Simulation settings}
Throughout the simulations, we fix the following settings. The optimization problems are formulated using CVX \cite{cvx,gb08} and solved using MOSEK \cite{mosek} and SDPT3 \cite{toh1999sdpt3}.
The SDPs are solved through Algorithm \ref{alg:solve_sdp}. For step \ref{step:cost_improve}, the cost of the rank-1 solution is improved using rank minimization (\ref{sec:gradient-based-rank-min}), low-rank channel (Problem \ref{prob:low_rank_channel}), and scheduling (\ref{eq:scheduling rank constraint}) interchangeably, following the fashion: Scheduling\textrightarrow rank minimization\textrightarrow Low rank channel\textrightarrow rank minimization\textrightarrow repeat. For the scheduling process, we use the following sequence of tolerances.
\[
\{\sigma^k|\sigma^k=\max(10^{-5},1-(1+e^{(25-k)/5})^{-1})\}
\]
This scheduling curve starts with a large tolerance that decreases rapidly; in practice, a similarly shaped linear curve yields comparable performance. Tuning of the parameters for the curve shape may result in different behaviors.

The iteration limits for each of the scheduling and low-rank channel phases are set as $1000$ and $200$, respectively. The sequence of scheduling and the low-rank channel is allowed to repeat once. 

Along the iterations, we measure the following values for the quality of solutions.
\begin{itemize}
    \item $\bar{\mR}_{a}:=\norm{\hat{\mR}_a\mR_a\transpose-\mI}$ is the error of a rotation $\mR_a$ from its ground truth value $\hat{\mR}_a$.
    \item $\bar{\vt}_a:=\norm{\vt_a-\hat{\vt}_a}$ is the error of a translation $\vt_a$ compared to its ground truth $\hat{\vt}_a$.
    \item The eigenvalue gap (EG) defined by $\bar{\lambda}-\min( \lambda_1(\mY))$, representing the proximity of the rank-1 property.
    \item The duality gap (DG) defined by $f^\star-d^\star$, used as a certificate for global optimality.
    \item A solution is marked \textit{success} when any of the rotational estimation error $\bar{\mR}\geq 0.1$.
\end{itemize}

\begin{figure}[ht]
    \vspace{5pt}
    \centering
    \subfloat[$\lambda_1(\mY^k_i),\lambda_1(\mY^k_{\tau,j})$]{\resizebox{0.49\textwidth}{!}{\includegraphics[width=0.95\linewidth]{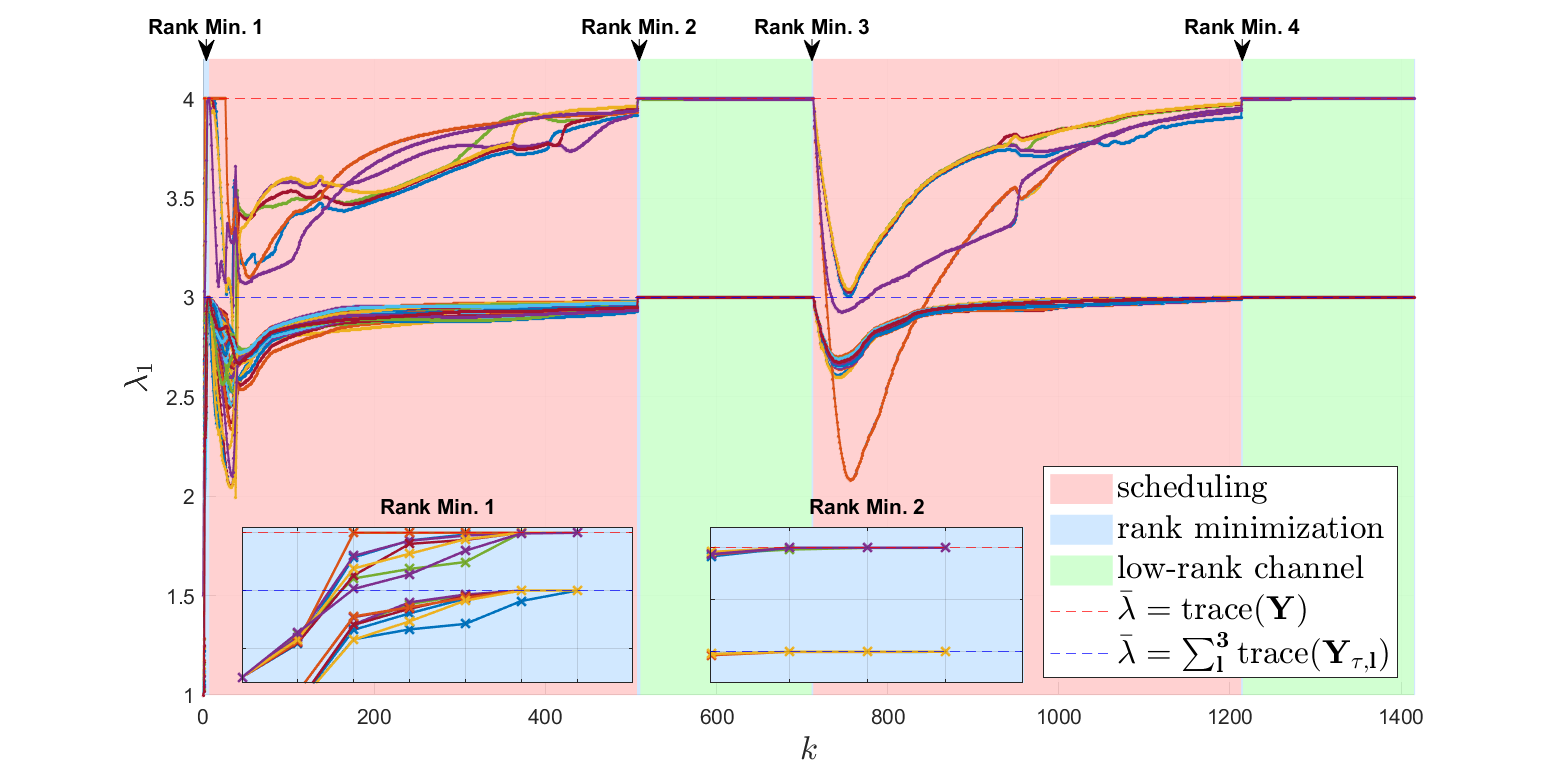}\label{fig:rank-iter}}}

    \subfloat[$f(\mY^k,\mY^k_\tau)$]{\resizebox{0.49\textwidth}{!}{\includegraphics[width=0.95\linewidth]{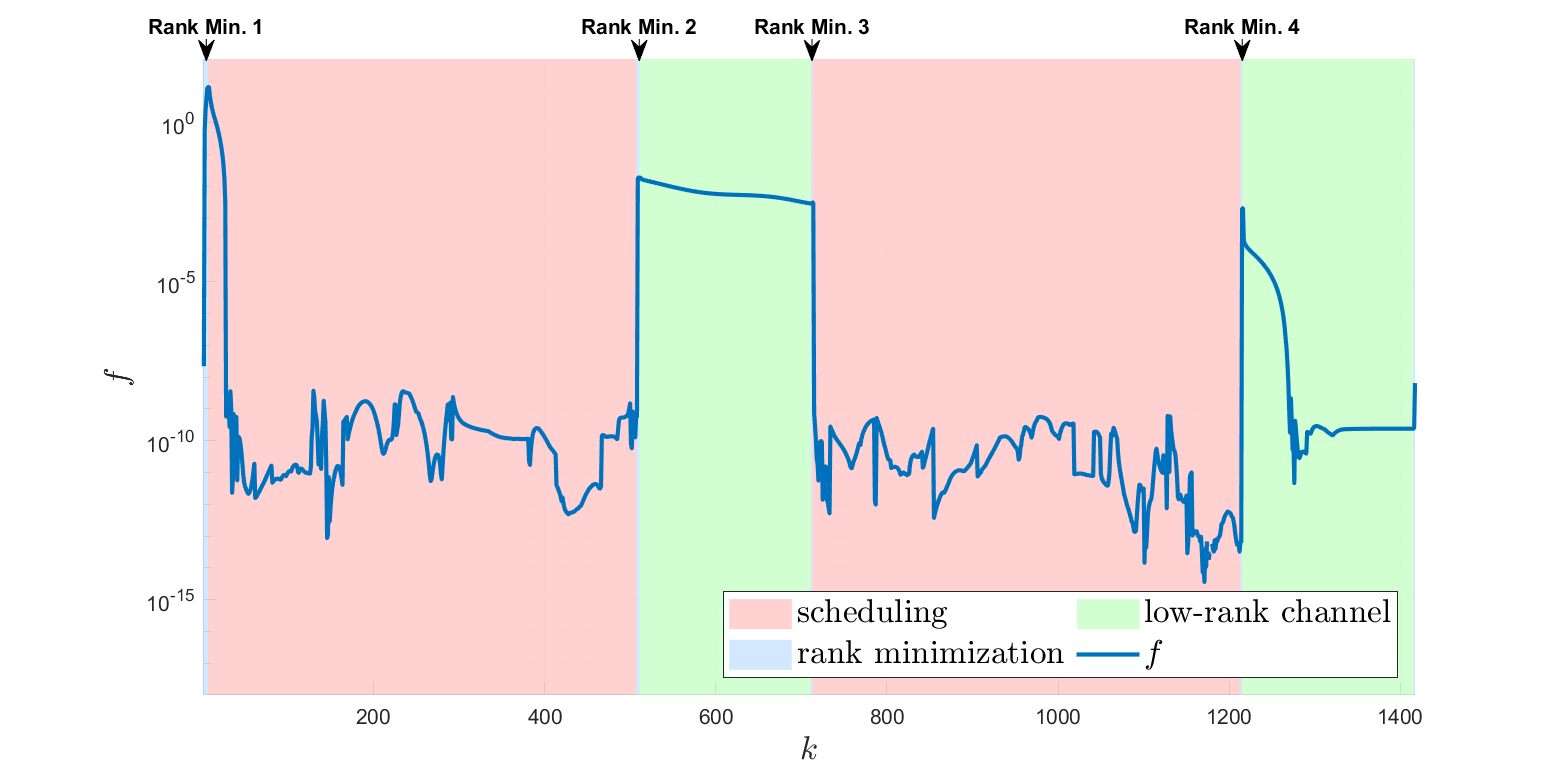}\label{fig:cost-iter}}}
    \caption{Over the iterations of Algorithm \ref{alg:solve_sdp}, the rank is minimized in the rank minimization and scheduling phases, while the cost is reduced during the scheduling and the low-rank channel phases.}
    \label{fig:rank-cost-iters}
\end{figure}

\begin{table*}[ht]
\centering
\vspace{3pt}
\resizebox{0.78\textwidth}{!}{
\begin{tabular}{cccccccccccc}
Problem & $m$ & $n$ & Noise Lv. $^\dag$ & $\bar{\mR}_c$ & $\bar{\vt}_c$ & EG & DG & Cost & Time & Iterations & \makecell{Success/\\Total Runs} \\ 
\hline 
  \multirow{5}{*}{PnP}& - & 10 & none & 3.44e-05 & 1.05e-04 & 2.70e-05 & 6.13e-09 & 6.37e-10 & 3.89e+02 & 1.01e+03 & 19/20 \\ 
  & - & 5 & none & 1.83e-05 & 6.11e-05 & 9.24e-05 & 4.22e-09 & 4.21e-09 & 2.97e+02 & 1.01e+03 & 20/20 \\ 
  & - & 10 & low & 7.24e-03 & 2.38e-02 & 3.93e-05 & 1.22e-04 & 1.22e-04 & 4.39e+02 & 1.10e+03 & 19/20 \\ 
  & - & 5 & low & 5.64e-03 & 1.99e-02 & 4.35e-06 & 5.40e-06 & 5.40e-06 & 3.02e+02 & 1.01e+03 & 20/20 \\ 
  & - & 10 & high & 6.19e-03 & 2.02e-02 & 5.40e-05 & 1.08e-04 & 1.11e-04 & 8.48e+02 & 1.93e+03 & 20/20 \\ 
  \hline
  \multirow{5}{*}{Hand-Eye}& 6 & 9 & none & 1.82e-03 & 1.03e-03 & 1.04e-06 & 6.90e-07 & 6.88e-07 & 1.24e+03 & 1.22e+03 & 7/10 \\ 
  & 6 & 9 & low & 5.32e-03 & 2.95e-03 & -3.52e-06 & 1.85e-04 & 8.66e-04 & 2.53e+03 & 2.40e+03 & 7/10 \\ 
  & 9 & 9 & none & 2.08e-03 & 1.13e-03 & * & 1.04e-06 & * & 2.38e+03 & 1.23e+03 & 9/10 \\ 
  & 9 & 9 & low & 1.48e-02 & 1.07e-02 & -1.37e-05 & 2.02e-03 & 2.68e-03 & 4.63e+03 & 2.34e+03 & 5/5 \\ 
  & 9 & 9 & high & 1.27e-02 & 8.23e-03 & -1.41e-05 & 2.11e-03 & 2.76e-03 & 4.61e+03 & 2.32e+03 & 5/5 \\ 
\hline 
\end{tabular}
}
\\[1ex]
\scriptsize{\dag $\text{ none: } \hat{e}_p=0, \quad \text{low: } \hat{e}_p=2, \quad \text{high: }\hat{e}_p=5$    $\qquad\mid\qquad$   *:data pending}
\caption{Average Estimation results of the PnP and hand-eye calibration problem}
\label{tab:pnp}
\end{table*}

\textit{\textbf{PnP estimation and Hand-eye calibration.}} We first evaluate the proposed method on two problems: camera pose estimation and hand–eye calibration. For camera pose estimation, a set of $n$ target points and a camera pose are randomly generated, and the corresponding images of the targets are obtained. To simulate measurement noise, the image features are perturbed on the image plane in both $x$ and $y$ directions by a random displacement bounded by $\hat{e}_p$ (pixels). 
For hand–eye calibration, the data are generated in the same manner, except the camera is mounted on a robot, and measurements are taken over $m$ distinct configurations.

\textit{\textbf{Dual-robot Calibration.}}
To evaluate our method on the dual-robot calibration example, we simulate random motions of two robots: a \textit{Rethink Robotics Sawyer} with a hand-mounted camera and a \textit{Kuka Iiwa 14} rigidly holding a calibration target. The robot bases are fixed to the ground. For each selected pair of configurations, the camera measurements are generated from the relative transformation between the calibration target and the camera, after which noise is added to the measured transformations $\mA$, $\mB$, and $\mC$.

For every problem, a TCSDP is constructed using the kinematics problems of the virtual robots. Then, the problem is solved through the proposed method. In the end, the transformations are recovered from the fixed-trace matrices.

\vspace{-5pt}
\subsection{Results}
\begin{table*}[ht]
\centering
\resizebox{0.98\textwidth}{!}{
\begin{tabular}{cccccccccccccc}
\# Meas. & Noise Lv.${{}^{\dag}}$ & $\bar{\mR}_{x}$ & $\bar{\mR}_{y}$ & $\bar{\mR}_{z}$ & $\bar{\vt}_{x}$ & $\bar{\vt}_y$ & $\bar{\vt}_z$ & EG & DG & Cost & CPU Time${}^{\dag\dag}$ & Iterations & \makecell{Success/\\Total Runs} \\ 
\hline 
6 & none & 5.63e-03 & 2.39e-03 & 7.60e-03 & 4.57e-02 & 1.02e-02 & 3.94e-02 & 2.65e-06 & * & 2.42e-04 & 1.74e+04 & 1.59e+03 & 15/20 \\ 
6 & medium & 9.89e-03 & 1.00e-02 & 1.22e-02 & 4.57e-02 & 1.28e-02 & 5.12e-02 & 6.59e-07 & 4.04e-04 & 5.40e-04 & 1.62e+04 & 1.64e+03 & 16/20 \\ 
9 & none & 6.19e-05 & 2.69e-05 & 4.75e-05 & 4.85e-05 & 6.61e-05 & 4.79e-05 & 7.32e-07 & 7.57e-07 & 7.43e-09 & 1.83e+04 & 1.21e+03 & 19/20 \\ 
9 & low & 2.03e-03 & 1.22e-03 & 2.27e-03 & 2.25e-03 & 1.18e-03 & 2.57e-03 & 1.32e-06 & 1.66e-05 & 1.58e-05 & 1.87e+04 & 1.21e+03 & 9/10 \\ 
9 & medium & 5.04e-03 & 2.17e-03 & 5.03e-03 & 2.89e-03 & 2.44e-03 & 1.77e-03 & 7.33e-07 & 9.98e-05 & 1.16e-04 & 5.01e+03 & 1.59e+03 & 9/10 \\ 
9 & high & 1.28e-02 & 5.01e-03 & 1.24e-02 & 8.90e-03 & 8.50e-03 & 1.29e-02 & 7.10e-07 & 8.86e-04 & 1.31e-03 & 3.96e+04 & 2.42e+03 & 17/20 \\ 
\hline 
\end{tabular}
}
\\[1ex]
{\scriptsize
*: data pending\\
\dag $\text{ low: } (\theta,l)=(0.1\degree,1e-4), \quad \text{medium: } (\theta,l)=(0.3\degree,3e-4), \quad \text{high: }(\theta,l)=(0.8\degree,8e-4)$. Noise added by $\tilde{\mR}=\mR\mathrm{rot}(\vv,\mathrm{rand}(0,\theta)), \tilde{\vt} = \vt+\mathrm{rand}(-l,l).$ \\
\dag\dag The CPU computation time measured for all CPU cores. Real clock time is approximately CPU time divided by the number of CPU cores.
}
\caption{Average Calibration results of the dual robot system}
\label{tab:results}
\end{table*}

A solution process for the dual-robot calibration example is shown in Fig. \ref{fig:rank-cost-iters}, where the largest eigenvalues of the fixed-trace matrices are plotted in Fig. \ref{fig:rank-iter} and the cost variation in Fig. \ref{fig:cost-iter}. The results show that the low-rank channels reduce cost while keeping $\lambda_1$ close to $\bar{\lambda}$. During scheduling, the cost decreases as $\lambda_1$ first drops and then rises toward $\bar{\lambda}$ as the tolerance tightens. After each phase, rank minimization projects solutions onto the rank-1 set, typically causing a cost increase.

Tables \ref{tab:pnp} and \ref{tab:results} report performance under varying measurement noise, showing higher average error and higher final cost with higher noise. In most cases, the solver converges to low-cost solutions; however, in a small fraction of cases, it fails to reach rank-1 low-cost solutions within the iteration limit, likely due to numerical issues or insufficient iterations. In successful cases, the eigenvalue gap is reduced under tolerance, showing the solutions are close to rank-1. In addition, a small average duality gap is observed, certifying global optimality. 

\vspace{-5pt}
\section{Conclusion}
We introduce the TCSDP, which enables gradient-based low-rank projection and optimality certification. A tailored solver is developed to obtain low-rank, low-cost solutions, with fixed-trace moment matrices designed for rotations and translations. Rank-1 solutions of these matrices allow exact recovery of the original manifold variables. We also present SP robot, a modular tool for modeling robotics problems via virtual robot kinematics, and demonstrate the full modeling–solving–certification pipeline on estimation and calibration tasks.

Future work includes accelerating the rank-minimization and cost-reduction procedure, as well as extending TCSDP to a broader range of robotics applications benefiting from convex optimization.

\ifshowfullcontents
\appendix
\subsection{Dual problem derivation}

\paragraph{Unified primal problem.}
We rewrite Problem \ref{prob:sdp-relax} as
\begin{subequations}
    \begin{align}
\min_{\{\mathbf{Y}_i\succeq \mathbf{0}\}} \quad
& f(\mathbf{Y}) \;=\; \vy\transpose\mathbf{Q}\,\vy \;+\; \mathbf{c}\transpose\vy \;+\; \text{constant} \label{eq:primal_quad_obj}\\[1ex]
\text{s.t.}\quad
& \mathcal{A}(\mathbf{Y}) \;=\; \mathbf{b}, \qquad \mathbf{Y}_i \succeq \mathbf{0}\ \ \forall i\in\{1,\dots,n\}. \label{eq:primal_constraints}
\end{align}
\end{subequations}
Here $\mathcal{A}$ is a given linear operator acting on the block tuple $\mathbf{Y}$, and $\mathbf{b}$ is given. This includes all of the linear (including trace) constraints in Problem \ref{prob:sdp-relax}.

\paragraph{Epigraph reformulation (linear objective + LMI).}
Let $\mathbf{Q}=\mathbf{L}\transpose\mathbf{L}$ be a Cholesky (or any) factorization and introduce an epigraph
variable $t\in\mathbb{R}$. Since
$\vy\transpose\mathbf{Q}\,\vy = \|\mathbf{L}\vy\|_2^2$,
the inequality $t \ge \vy\transpose\mathbf{Q}\,\vy$ admits the Schur-complement LMI
\[
\begin{bmatrix}
t & (\mathbf{L}\vy)\transpose \\
\mathbf{L}\vy & \mathbf{I}
\end{bmatrix}\succeq \mathbf{0},
\]
where $\mathbf{I}$ is the identity of conformable size.
Dropping the additive constant in \eqref{eq:primal_quad_obj} (it does not affect optimal $\mathbf{Y}$), the problem is equivalent to
\begin{subequations}
    \begin{align}
\min_{\{\mathbf{Y}_i\succeq \mathbf{0}\},\, t} \quad
& t \;+\; \mathbf{c}\transpose\vy \label{eq:primal_epi_obj}\\[0.5ex]
\text{s.t.}\quad
& \mathcal{A}(\mathbf{Y}) \;=\; \mathbf{b}, \qquad \mathbf{Y}_i \succeq \mathbf{0}\ \ \forall i, \label{eq:primal_epi_constr1}\\[0.5ex]
& \begin{bmatrix}
t & (\mathbf{L}\vy)\transpose \\
\mathbf{L}\vy & \mathbf{I}
\end{bmatrix}\succeq \mathbf{0}. \label{eq:primal_epi_constr2}
\end{align}
\end{subequations}
This is a standard-form SDP with a linear objective and affine LMI constraints in $(\mathbf{Y},t)$.

\paragraph{Dual derivation.}
Introduce dual variables:
\begin{itemize}
\item $\boldsymbol{\rho}$ for the equality constraints $\mathcal{A}(\mathbf{Y})=\mathbf{b}$,
\item $\{\mathbf{S}_i\succeq \mathbf{0}\}$ for $\mathbf{Y}_i\succeq \mathbf{0}$,
\item $\mathbf{Z}\succeq \mathbf{0}$ for the LMI \eqref{eq:primal_epi_constr2}, partitioned as
$\displaystyle \mathbf{Z}=\begin{bmatrix}\alpha & \mathbf{z}\transpose\\[0.2ex]\mathbf{z} & \mathbf{Z}_{22}\end{bmatrix}$,
where $\alpha\in\mathbb{R}$, $\mathbf{z}\in\mathbb{R}^{r}$ and $\mathbf{Z}_{22}\in\mathbb{S}^r$ with $r$ the row dimension of $\mathbf{L}$.
\end{itemize}
The Lagrangian of \eqref{eq:primal_epi_obj}–\eqref{eq:primal_epi_constr2} is
{\scriptsize\begin{equation}
\begin{aligned}
    \mathcal{L}(\mathbf{Y},t;\boldsymbol{\rho},\{\mathbf{S}_i\},\mathbf{Z})
= &\big(t+\mathbf{c}\transpose\vy\big)
+ \boldsymbol{\rho}\transpose\!\big(\mathbf{b}-\mathcal{A}(\mathbf{Y})\big)
{-} \sum_{i=1}^n \ip{\mathbf{S}_i}{\mathbf{Y}_i}\\
&- \ip{\mathbf{Z}}{\begin{bmatrix} t & (\mathbf{L}\vy)\transpose \\ \mathbf{L}\vy & \mathbf{I}\end{bmatrix}}.
\end{aligned}
\end{equation}}

Using the block structure of $\mathbf{Z}$,
\[
\ip{\mathbf{Z}}{\begin{bmatrix} t & (\mathbf{L}\vy)\transpose \\ \mathbf{L}\vy & \mathbf{I}\end{bmatrix}}
= \alpha\, t \;+\; 2\,\mathbf{z}\transpose\mathbf{L}\vy \;+\; \operatorname{tr}(\mathbf{Z}_{22}).
\]
Hence
{\scriptsize
\begin{equation}
    \begin{aligned}
        \mathcal{L}
= &(1-\alpha)\,t \;+\; \big(\mathbf{c}-2\mathbf{L}\transpose\mathbf{z}\big)\transpose\vy
\;+\; \boldsymbol{\rho}\transpose\mathbf{b}
\;-\; \ip{\mathcal{A}^*(\boldsymbol{\rho})}{\mathbf{Y}}\\
&\;-\; \sum_{i=1}^n \ip{\mathbf{S}_i}{\mathbf{Y}_i}
\;-\; \operatorname{tr}(\mathbf{Z}_{22}),
    \end{aligned}
\end{equation}
}
where $\mathcal{A}^*$ is the adjoint of $\mathcal{A}$ and
$\ip{\mathcal{A}^*(\boldsymbol{\rho})}{\mathbf{Y}} \coloneqq \sum_{i=1}^n \ip{\big(\mathcal{A}^*(\boldsymbol{\rho})\big)_i}{\mathbf{Y}_i}$.

For the dual function $g$, take the infimum over $(\mathbf{Y},t)$.
This is finite iff the coefficients of $t$ and of each $\mathbf{Y}_i$ vanish:
\begin{align}
&\alpha = 1, \label{eq:dual_cond_alpha}\\[0.25ex]
&\mathcal{V}^*\!\big(\mathbf{c}-2\mathbf{L}\transpose\mathbf{z}\big)\;-\;\mathcal{A}^*(\boldsymbol{\rho})\;-\;\mathbf{S}
\;=\; \mathbf{0}, \label{eq:dual_stationarity}
\end{align}
where $\mathbf{S}:=\{\mathbf{S}_i\}_{i=1}^n$, and $\mathcal{V}:\mathbf{Y}\mapsto \vy=\vecop(\mathbf{Y})$ and $\mathcal{V}^*$ is its adjoint, i.e.,
$\ip{\mathcal{V}^*(\mathbf{r})}{\mathbf{Y}} = \mathbf{r}\transpose\vecop(\mathbf{Y})$ for any $\mathbf{r}\in\mathbb{R}^{N}$.
Under \eqref{eq:dual_cond_alpha},\eqref{eq:dual_stationarity}, the infimum equals
$\boldsymbol{\rho}\transpose\mathbf{b} - \operatorname{tr}(\mathbf{Z}_{22})=\boldsymbol{\rho}\transpose\mathbf{b} - \operatorname{tr}(\mathbf{Z})+1$.
Therefore, the dual SDP is exactly Problem \ref{prob:dual}


\fi




\bibliographystyle{ieee}

\bibliography{main}

\end{document}